\documentclass{article}
\usepackage[utf8]{inputenc}
\usepackage{xcolor}
\usepackage{colortbl}

\definecolor{better}{HTML}{D5E8D4} 
\definecolor{worse}{HTML}{F8CECC}  

\usepackage{microtype}
\usepackage{graphicx}
\usepackage{subcaption}
\usepackage{booktabs} 
\usepackage{multirow}
\usepackage{hyperref}
\usepackage{enumitem}

\newcommand{\sig}[1]{#1\rlap{$^*$}}

\usepackage[preprint]{icml2026}

\usepackage{amsmath}
\usepackage{amssymb}
\usepackage{mathtools}
\usepackage{amsthm}

\usepackage[capitalize,noabbrev]{cleveref}

\theoremstyle{plain}

\theoremstyle{definition}

\theoremstyle{remark}

\usepackage[textsize=tiny]{todonotes}

\icmltitlerunning{Order-Aware Test-Time Adaptation: Leveraging Temporal Dynamics for Robust Streaming Inference}

\begin{document}

\twocolumn[
  \icmltitle{Order-Aware Test-Time Adaptation: \\
    Leveraging Temporal Dynamics for Robust Streaming Inference}




  \begin{icmlauthorlist}
    \icmlauthor{Young Kyung Kim}{princeton,duke}
    \icmlauthor{Oded Schlesinger}{princeton,duke}
    \icmlauthor{Qiangqiang Wu}{princeton,cityu}
    \icmlauthor{J. Mat\'ias Di Martino}{duke,catolica}
    \icmlauthor{Guillermo Sapiro}{princeton,apple}
  \end{icmlauthorlist}

  \icmlaffiliation{princeton}{Department of Electrical and Computer Engineering, Princeton University, Princeton, USA}
  \icmlaffiliation{duke}{Department of Electrical and Computer Engineering, Duke University, Durham, USA}
  \icmlaffiliation{cityu}{Department of Computer Science, City University of Hong Kong, Hong Kong, China}
  \icmlaffiliation{catolica}{Department of Informatics and Computer Science, Universidad Católica del Uruguay, Montevideo, Uruguay}
  \icmlaffiliation{apple}{Apple, Cupertino, USA}

  \icmlcorrespondingauthor{Young Kyung Kim}{yk4491@princeton.edu}
  \icmlcorrespondingauthor{Guillermo Sapiro}{guillermos@princeton.edu}

  \icmlkeywords{Machine Learning, ICML}

  \vskip 0.3in
]



\printAffiliationsAndNotice{}  

\begin{abstract}
Test-Time Adaptation (TTA) enables pre-trained models to adjust to distribution shift by learning from unlabeled test-time streams. However, existing methods typically treat these streams as independent samples, overlooking the supervisory signal inherent in temporal dynamics. To address this, we introduce \textbf{O}rder-\textbf{A}ware \textbf{T}est-\textbf{T}ime \textbf{A}daptation \textbf{(OATTA)}. We formulate test-time adaptation as a gradient-free recursive Bayesian estimation task, using a learned dynamic transition matrix as a temporal prior to refine the base model's predictions. To ensure performance in weakly structured streams, we introduce a \textit{likelihood-ratio gate} (LLR) that reverts to the base predictor when temporal evidence is absent. OATTA is a lightweight, model-agnostic module that incurs negligible computational overhead. Extensive experiments across image classification, wearable and physiological signal analysis, and language sentiment analysis demonstrate its universality; OATTA consistently boosts established baselines, improving accuracy by up to 6.35\%. Our findings establish that modeling temporal dynamics provides a critical, orthogonal signal beyond standard order-agnostic TTA approaches.
\end{abstract}


\section{Introduction}

\begin{figure}[t]
    \centering
    \includegraphics[width=0.9\linewidth]{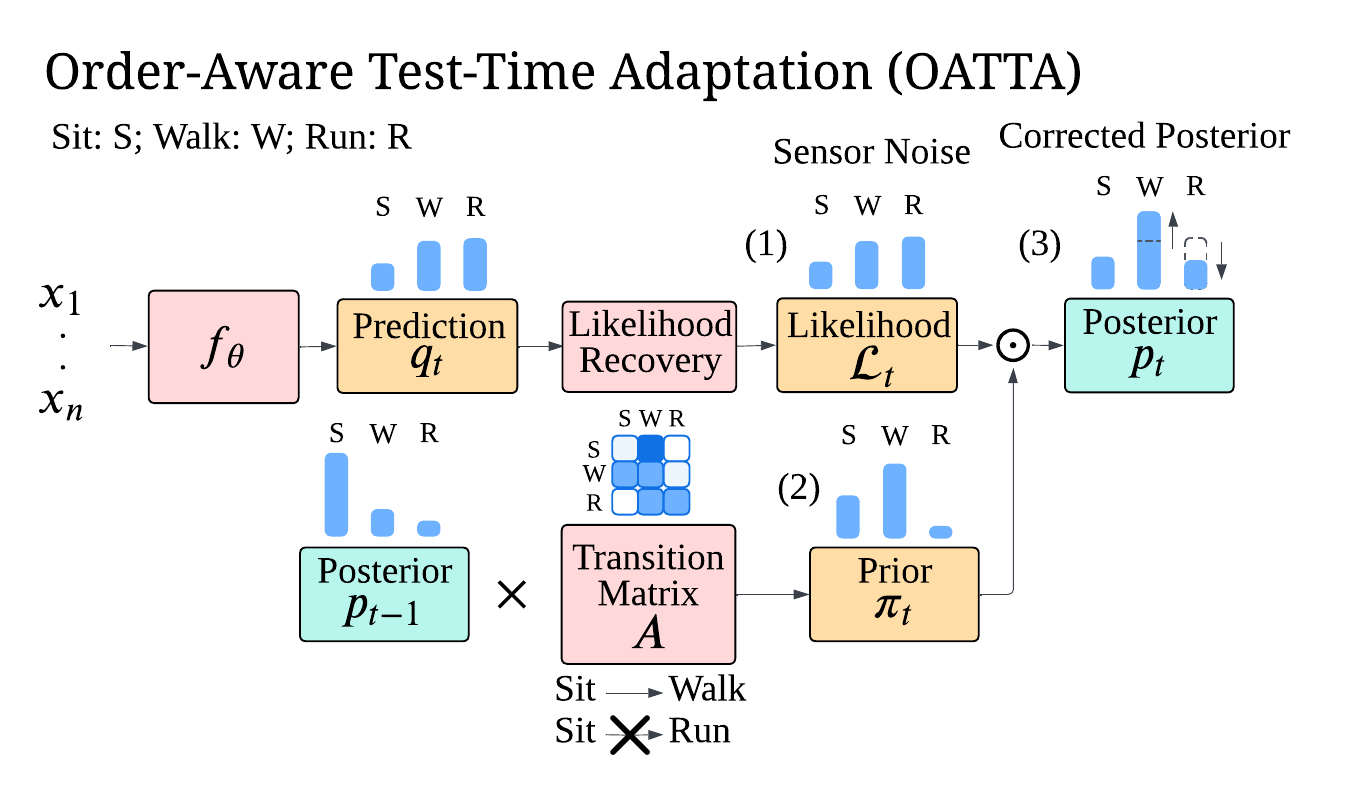}
    \caption{\textbf{Illustration of Order-Aware Test-Time Adaptation (OATTA) using a human activity recognition example.} Unlike order-agnostic TTA, OATTA leverages temporal structure through recursive Bayesian estimation. (1) The problem (sensor ambiguity): As shown in the ``Sensor Noise'' bar chart, the backbone's raw prediction ($q_t$) can be ambiguous (e.g., confusing ``Walk'' with ``Run''). We convert $q_t$ to a likelihood ($\mathcal{L}_t$) using a uniform prior ($\mathcal{U}$), since source statistics are unavailable. (2) The solution (temporal logic): This likelihood is fused with a temporal prior ($\pi_t$), obtained by projecting the previous posterior ($p_{t-1}$) through the dynamic transition matrix ($A$). (3) Result: In this example, although the sensor is uncertain, OATTA promotes ``Walk'' over ``Run.'' This occurs because $A$ has learned that the transition ``Sit~$\to$~Run'' is physically improbable, effectively using temporal context to filter out the sensor noise.}
    \label{fig:system_overview}
\end{figure}

Test-Time Adaptation (TTA) is a standard tool for handling distribution shift at deployment time, updating a pre-trained model using only unlabeled test data~\citep{sun2020test,wang2020tent,zhang2022memo}. Existing methods primarily focus on adapting model representations to the current test distribution. While effective, this perspective treats the test stream as a collection of i.i.d.\ samples. Consequently, it overlooks a critical source of information available in many real-world deployments, namely the temporal dynamics linking consecutive observations. Examples include human activity streams, wildlife camera-trap sequences (e.g., herd and predator–prey patterns), and social-media timelines where sentiment persists across posts. By ignoring this temporal context, the model loses a critical signal for resolving ambiguity and uncertainty, thereby compromising its performance and robustness.

We posit that explicitly modeling these temporal transition dynamics offers a powerful, complementary mechanism for online adaptation. Most TTA methods primarily address marginal distribution shift in the inputs or learned representations (i.e., changes in $p(x)$ or $p(z)$). In contrast, our approach leverages sequential dependence by modeling class-transition dynamics (i.e., $p(y_t \mid y_{t-1})$) to regularize predictions. This distinction enables the model to adapt to the unique temporal structure that governs the current test-time-adaptation stream. For instance, an office worker and a professional athlete may both generate ``walk'' signals that standard TTA can align. However, their transition dynamics differ fundamentally: the athlete may transition ``walk $\to$ run'', while the office worker transitions ``walk $\to$ sit''. Neither static models nor standard TTA methods can capture these user-specific routines. By learning the transition matrix online, our method dynamically tailors the predictive prior to the stream's specific dynamics, improving robustness without requiring target-specific training data.

We propose a model-agnostic framework for \textbf{O}rder-\textbf{A}ware \textbf{T}est-\textbf{T}ime \textbf{A}daptation \textbf{(OATTA)}, as illustrated in Figure~\ref{fig:system_overview}. We formulate the problem as a gradient-free \textit{recursive Bayesian estimation} task, maintaining a dynamic transition matrix that serves as a temporal prior and is fused with the base model's predictions at each step. For reliable deployment when temporal structure is weak or absent, we introduce an optional \textit{likelihood-ratio gate} (LLR) that falls back to the base predictor and only permits adaptation when the temporal prior yields consistent evidence beyond an order-agnostic baseline.


Our main contributions are as follows:
\begin{itemize}[leftmargin=*, noitemsep]
    \item \textbf{Problem Formulation:} To the best of our knowledge, we are the first to exploit the latent class-transition matrix for general TTA to address temporal information. We demonstrate that modeling dynamics provides a critical source of information orthogonal to traditional domain alignment.
    \item \textbf{Algorithm:} We propose a lightweight Bayesian filtering wrapper that learns stream-specific transition dynamics online. Our method is model-agnostic, works on top of TTA baselines (e.g., Tent and MEMO), and incurs negligible computational overhead.
    \item \textbf{Empirical Validation:} We evaluate OATTA on controlled benchmarks and real-world datasets spanning image classification, wearable and physiological signal analysis, and language sentiment analysis. Our results demonstrate that OATTA is generalizable, improving performance across modalities and boosting established TTA baselines. Crucially, we validate the framework's adaptation in recovering from non-stationary regime shifts. 
\end{itemize}

\section{Related Work}

\textbf{Domain Adaptation (DA).} 
DA mitigates performance degradation under distribution shift between a labeled source domain and an unlabeled target domain~\citep{pan2009survey}. Discrepancy-based methods such as DDC~\citep{tzeng2014deep} and DAN~\citep{long2015learning} minimize maximum mean discrepancy, while JAN~\citep{long2017deep} aligns joint feature--prediction distributions. Adversarial approaches learn invariant representations, e.g., DANN~\citep{ganin2016domain} with GRL and extensions such as ADDA~\citep{tzeng2017adversarial} and CDAN~\citep{long2018conditional}; generative variants (CoGAN~\citep{liu2016coupled}, PixelDA~\citep{bousmalis2017unsupervised}) synthesize source-like target images. These methods necessitate concurrent access to source and target data, a constraint that TTA and methods like the one proposed herein seek to eliminate.

\textbf{Test-Time Adaptation (TTA).} 
TTA adapts a pre-trained model to unlabeled target data during inference, obviating the need to access source data at test time~\citep{liang2020we, sun2020test}. Most existing approaches are optimization-based, e.g., TTT~\citep{sun2020test} and Tent~\citep{wang2020tent} update model parameters via self-supervision or entropy minimization, while CoTTA~\citep{wang2022continual} and EATA~\citep{niu2022efficient} address catastrophic forgetting in streaming settings. Gradient-free approaches have also emerged to avoid the cost of backpropagation at test time, e.g., SHOT~\citep{liang2020we} and T3A~\citep{iwasawa2021test} utilize pseudo-labeling to adjust classifier prototypes, while LAME~\citep{boudiaf2022parameter} refines predictions via manifold smoothness constraints. Among instance-level methods, MEMO~\citep{zhang2022memo} enforces consistency across augmentations but treats each sample in isolation. NOTE~\citep{gong2022note} treats temporal correlation as a bias to mitigate, whereas OATTA treats it as a form of supervision. By explicitly modeling transition dynamics, OATTA distinguishes itself from prior methods, which typically ignore the temporal dependencies in sequential data.


\textbf{Bayesian Filtering \& Deep State-Space Models.} 
Classical filtering frameworks, such as Kalman filters~\citep{kalman1960new} and Hidden Markov models~\citep{rabiner2002tutorial}, provide a rigorous basis for sequential estimation. Recent advances have integrated these principles with deep learning to model complex non-linear dynamics: Differentiable KFs~\citep{haarnoja2016backprop} combine neural networks with probabilistic graphical models, while modern state-space models (SSMs) like S4~\citep{gu2021efficiently} and Mamba~\citep{gu2024mamba} leverage continuous-time control theory for efficient sequence modeling. In video recognition, methods like TeSTra~\citep{zhao2022real} apply temporal smoothing kernels to stabilize frame-wise predictions. However, these approaches rely on extensive offline training to learn the underlying dynamics model. In contrast, the here proposed OATTA adapts the filtering paradigm for the online test-time setting. Instead of updating on pre-trained weights, OATTA uses recursive estimation to update a dynamic transition matrix in real-time. 


\section{Methodology}

We propose Order-Aware Test-Time Adaptation (OATTA), a Bayesian framework for robust streaming inference. Unlike optimization-based TTA methods that adapt model parameters $\theta$ to minimize prediction uncertainty, OATTA is a training-free post-processing method. It operates strictly in the output probability space, maintaining a dynamic transition prior over classes to regularize predictions without modifying the model weights.

\subsection{Preliminaries}
Consider a data stream $\mathcal{S} = \{x_1, x_2, \dots, x_T\}$ arriving sequentially. At each step $t$, a base model $f_\theta$ produces a probability vector $q_t = f_\theta(x_t) \in [0,1]^K$, where $K$ is the number of classes, with $\sum_{k=1}^K q_{t}(k) = 1$. Interpreting $q_t$ as an estimate of the posterior, $q_t(k) \approx p(y{=}k \mid x_t)$, Bayes’ rule implies a likelihood proportional to $\mathcal{L}_t(k) \propto \frac{q_t(k)}{\rho(k)}$, where $\rho(k)=p(y{=}k)$ is the class prior. We default to a uniform class prior, i.e., $\rho=\mathcal{U}$, since source statistics are typically unavailable in standard TTA; however, we allow substituting source metadata if accessible to explicitly mitigate label shift between the source and target domains. An analysis of source-metadata substitution is provided in Appendix~\ref{app:lds_robustness}. Here, $\mathcal{U}$ denotes the uniform distribution over $K$ classes, i.e., $\mathcal{U}(k)=\frac{1}{K}$.


Standard TTA methods typically process data as an i.i.d.\ stream, ignoring sequential dependencies. In contrast, we assume the stream is generated by a Hidden Markov model characterized by an unknown, potentially non-stationary transition matrix $A^*$. The joint distribution factorizes as
\begin{equation}
    p(x_{1:T}, y_{1:T}) = p(y_1) \prod_{t=2}^T p(y_t | y_{t-1}; A^*) \prod_{t=1}^T p(x_t | y_t).
\end{equation}
Our objective is to recursively infer the posterior marginal $p(y_t | x_{1:t})$ while simultaneously estimating the time-varying transition matrix $A_t \approx A^*$.

\subsection{Recursive Bayesian Estimation}
We formulate the adaptation as a recursive Bayesian filter. We maintain a running estimate of the transition matrix $A_t \in \mathbb{R}^{K \times K}$. The filtering process consists of two steps performed at each time $t$:

\textbf{1. Prediction (time update).} 
Before observing the current input $x_t$, we compute a temporal prior $\pi_t$ by projecting the previous posterior $p_{t-1}$ forward through the learned dynamics,
\begin{equation}
    \pi_t = A_{t-1}^\top p_{t-1}.
\end{equation}
This step redistributes probability mass based on the learned temporal dynamics.

\textbf{2. Measurement (measurement update).} 
We fuse the temporal prior $\pi_t$ with the likelihood $\mathcal{L}_t$ via point-wise multiplication.
\begin{equation}
    p_t(k) = \frac{\mathcal{L}_t(k) \cdot \pi_t(k)}{\sum_{j=1}^K \mathcal{L}_t(j) \cdot \pi_t(j)}.
\end{equation}

\textbf{Mechanism of Action (contextual fusion).}
The update balances observational evidence $\mathcal{L}_t$ with contextual evidence $\pi_t$. In the limit where one source becomes uninformative (approaches a uniform distribution), the posterior automatically converges to the remaining informative source,
\begin{equation}
    p_t(k) \propto 
    \begin{cases} 
      \frac{1}{K} \cdot \pi_t(k) \implies p_t \approx \pi_t & \text{if } \mathcal{L}_t \to \mathcal{U}, \\
      \mathcal{L}_t(k) \cdot \frac{1}{K} \implies p_t \approx \mathcal{L}_t & \text{if } \pi_t \to \mathcal{U},
   \end{cases}
\end{equation}
\noindent This ensures the filter corrects predictions when the input is noisy but falls back to the base model when dynamics are uninformative.

    


\subsection{Online Transition Estimation}
The transition matrix $A_t$ is not fixed and must be learned online. We estimate $A_t$ by maintaining a running count matrix $C_t \in \mathbb{R}^{K \times K}$ initialized with uniform pseudocounts $\kappa$. The update rule is an exponential moving average (EMA),
\begin{equation}
    C_t = (1 - \gamma w_t) C_{t-1} + \gamma w_t (\underbrace{q_{t-1} \otimes q_t}_{\text{decoupled signal}}),
\end{equation}
where $\gamma$ is the forgetting rate, $w_t$ is the entropy gating factor (detailed below), and $A_t$ is obtained by row-normalizing $C_t$. 


\textbf{1. Decoupled Updates.} We update $C_t$ using the raw model predictions ($q_{t-1}, q_t$) rather than the filtered posteriors. This isolates the transition estimation from the filter's own corrections, preventing confirmation bias loops and ensuring $A_t$ remains grounded in the observed data.

\textbf{2. Entropy Gating ($w_t$).} We modulate the update magnitude based on the confidence of the current prediction using the Shannon entropy $H(q_t)$, \begin{equation} w_t = \exp(-H(q_t)/\tau). \end{equation} If the base model is uncertain (high entropy), $w_t \to 0$, effectively skipping the update to prevent learning from noisy samples. Algorithm~\ref{alg:oatta} summarizes the full OATTA procedure.

\subsection{Likelihood-Ratio Gate (LLR)}
\label{sec:llr_gate}

The Bayesian update above yields a temporally regularized posterior that is most beneficial when the stream contains genuine sequential structure. On nearly i.i.d.\ streams, however, an online transition prior can potentially overfit noise and incur performance degradation. To make OATTA conservative in such regimes, we introduce a likelihood-ratio gate (LLR) that tests whether exploiting temporal order improves predictive fit relative to an order-agnostic baseline that ignores sequence information. The gate then interpolates between the Bayes-filtered posterior and the base predictor.

Let $\pi_t$ denote the temporal prior from the prediction step (Equation~(3)), and let $\bar{\pi}_t$ denote an \emph{order-agnostic baseline prior} (e.g., a slowly varying estimate of class frequencies). We evaluate which prior better explains the current model output $q_t$ using a log-score difference,
\begin{equation}
\Delta_t
=\log\!\big(\langle q_t,\pi_t\rangle+\varepsilon\big)
-\log\!\big(\langle q_t,\bar{\pi}_t\rangle+\varepsilon\big),
\end{equation}
and aggregate evidence over a short window of length $W$,
\begin{equation}
\mathrm{LLR}_t=\frac{1}{W}\sum_{i=0}^{W-1}\Delta_{t-i}.
\end{equation}
We convert this statistic into a mixing weight
\begin{equation}
\lambda_t=\sigma\!\left(\frac{\mathrm{LLR}_t-m}{\tau}\right),
\qquad \sigma(z)=\frac{1}{1+e^{-z}}.
\end{equation}
Here, $m$ is a margin that biases the gate toward the order-agnostic explanation unless the temporal prior provides sufficiently stronger evidence; increasing $m$ makes the method more conservative on i.i.d.\ streams. The parameter $\tau$ controls the sharpness of the transition; a smaller $\tau$ produces a sharper transition, while a larger $\tau$ yields smoother interpolation.

Finally, letting $p_t$ denote the Bayes-filtered posterior from Equation~(4), we form the gated prediction
\begin{equation}
\hat{p}_t=\lambda_t\,p_t+(1-\lambda_t)\,q_t.
\qquad
\end{equation}
When temporal structure is unsupported (small $\mathrm{LLR}_t$), $\lambda_t$ decreases, and the method falls back to $q_t$; when the temporal prior is consistently predictive (large $\mathrm{LLR}_t$), $\lambda_t$ increases and OATTA relies more heavily on the Bayes-filtered posterior. We provide the full specification of $\bar{\pi}_t$, the windowing strategy, and the complete gated algorithm in Appendix~\ref{app:llr_gate}.


\subsection{Computational Efficiency}
We now analyze the inference cost relative to a standard ResNet50 backbone; the analysis extends to other frontier models. Let $C_{\text{fwd}}$ denote the floating-point operations (FLOPs) count required for a single forward pass ($\approx 3.8$ GFLOPs)~\citep{he2016deep}. OATTA operates as a lightweight post-processing step on the output probability vector, incurring a fixed additive cost of $\mathcal{O}(K^2)$. For CIFAR-10 ($K=10$), this adds only $\approx 700$ FLOPs ($\approx$ 0.00002\% overhead). In stark contrast, gradient-based methods such as Tent require backpropagation to update model parameters; since the backward pass is approximately twice as expensive as the forward pass~\citep{kaplan2020scaling}, these methods essentially triple the inference computational budget per sample. See Appendix~\ref{app:complexity_proof} for detailed FLOPs analysis.

\begin{algorithm}[t]
   \caption{Order-Aware Test-Time Adaptation (OATTA)}
   \label{alg:oatta}
\begin{algorithmic}[1]
   \STATE {\bfseries Input:} Stream $\mathcal{S}$, Model $f_\theta$, Prior $\rho$ (default: uniform)
   \STATE {\bfseries Init:} $C_0 \leftarrow \kappa I$; \quad $A_0 \leftarrow \text{RowNorm}(C_0)$; \newline \quad $p_0, q_0 \leftarrow \mathbf{1}/K$
   \FOR{$t=1$ {\bfseries to} $T$}
       \STATE $q_t \leftarrow f_\theta(x_t)$; \quad $\pi_t \leftarrow A_{t}^\top p_{t-1}$ \hfill \textcolor{gray}{// Predict}
       \STATE $p_t \propto (q_t / \rho) \odot \pi_t$ \hfill \textcolor{gray}{// Bayes Update}
       
       \STATE $w_t \leftarrow e^{-H(q_t)/\tau}$ \hfill \textcolor{gray}{// Entropy Gate}
       \STATE $C_t \leftarrow (1 \!-\! \gamma w_t)C_{t-1} + \gamma w_t (q_{t-1} \otimes q_t)$ 
       \STATE $A_t \leftarrow \text{RowNorm}(C_t)$; \quad $p_{t-1} \leftarrow p_t; \quad q_{t-1} \leftarrow q_t$
   \ENDFOR
\end{algorithmic}
\end{algorithm}

\begin{table*}[t]
  \centering
  \caption{\textbf{Domain-centric view of universality.} Results are grouped by domain to highlight robustness across architectures and modalities. Bold indicates better performance between each baseline and its + Ours counterpart. $\Delta_{\text{Base}}$ denotes mean improvement relative to Base.}
  \label{combined_vertical_no_horiz}
  \setlength{\tabcolsep}{1.8pt}
  \resizebox{\textwidth}{!}{%
  \begin{sc}
    \begin{tabular}{l| cc cc cc cc cc cc ccc | c}
      \toprule
      & \multicolumn{6}{c}{\textbf{Image Domain}}
      & \multicolumn{6}{c}{\textbf{Sensor Domain}}
      & \multicolumn{3}{c}{\textbf{Language Domain}}
      & \multicolumn{1}{c}{} \\
      \cmidrule(lr){2-7} \cmidrule(lr){8-13} \cmidrule(lr){14-16} 
      & \multicolumn{2}{c}{ResNet50} & \multicolumn{2}{c}{ViT Base} & \multicolumn{2}{c}{ConvNeXt}
      & \multicolumn{2}{c}{CNN} & \multicolumn{2}{c}{TCN} & \multicolumn{2}{c}{1D ResNet18}
      & T-BERT & D-BERT & BERT
      & $\Delta_{\text{Base}}$ \\
      \cmidrule(lr){2-3} \cmidrule(lr){4-5} \cmidrule(lr){6-7}
      \cmidrule(lr){8-9} \cmidrule(lr){10-11} \cmidrule(lr){12-13}
      \cmidrule(lr){14-14} \cmidrule(lr){15-15} \cmidrule(lr){16-16}
      Method
      & CCT & UNSW & CCT & UNSW & CCT & UNSW
      & HARTH & Sleep & HARTH & Sleep & HARTH & Sleep
      & SENT & SENT & SENT
      &  \\
      \midrule

      Base
      & 54.30 & 92.24 & 53.17 & 91.70 & 58.88 & 92.33 & 83.82 & 75.97 & 82.78 & 76.16 & 84.58 & 74.05 & 84.25 & 85.25 & 85.84
      & -- \\
      \textbf{+ Ours}
      & \textbf{57.08} & \textbf{93.85} & \textbf{55.24} & \textbf{93.07} & \textbf{61.84} & \textbf{93.53} & \textbf{84.97} & \textbf{76.05} & \textbf{83.64} & \textbf{77.07} & \textbf{85.37} & \textbf{75.35} & \textbf{86.27} & \textbf{86.93} & \textbf{87.35}
      & \textbf{+1.49} \\
      \midrule

      MC-Dr
      & 54.32 & 92.39 & 53.18 & 91.70 & 58.85 & 92.33 & 83.98 & 76.02 & 82.79 & 76.30 & 84.60 & 74.09 & 84.43 & 85.21 & 85.89
      & +0.05 \\
      \textbf{+ Ours}
      & \textbf{57.09} & \textbf{93.82} & \textbf{55.24} & \textbf{93.07} & \textbf{61.84} & \textbf{93.53} & \textbf{85.02} & \textbf{76.13} & \textbf{83.73} & \textbf{77.07} & \textbf{85.39} & \textbf{75.51} & \textbf{86.33} & \textbf{86.89} & \textbf{87.38}
      & \textbf{+1.51} \\
      \midrule

      TTAug
      & 56.95 & 95.20 & 56.74 & 94.39 & 62.09 & 95.67 & 86.43 & \textbf{75.20} & 85.57 & 75.02 & 85.09 & 75.43 & 83.02 & 83.80 & 84.30
      & +1.31 \\
      \textbf{+ Ours}
      & \textbf{59.46} & \textbf{95.92} & \textbf{58.30} & \textbf{95.21} & \textbf{64.65} & \textbf{96.40} & \textbf{86.93} & 74.88 & \textbf{87.08} & \textbf{75.62} & \textbf{85.78} & \textbf{76.72} & \textbf{84.34} & \textbf{84.75} & \textbf{85.19}
      & \textbf{+2.39} \\
      \midrule

      Tent
      & \textbf{52.87} & \textbf{94.92} & 53.83 & 92.17 & 60.02 & 93.23 & 84.19 & 74.88 & 83.09 & 75.98 & 84.70 & 73.93 & 85.23 & 86.32 & 86.71
      & +0.45 \\
      \textbf{+ Ours}
      & 52.29 & 94.72 & \textbf{54.33} & \textbf{92.65} & \textbf{60.05} & \textbf{93.50} & \textbf{84.96} & \textbf{76.17} & \textbf{83.61} & \textbf{77.11} & \textbf{85.33} & \textbf{75.18} & \textbf{86.03} & \textbf{87.08} & \textbf{87.41}
      & \textbf{+1.01} \\
      \midrule

      MEMO
      & 54.49 & 93.91 & 54.62 & 92.01 & 61.10 & 93.88 & 83.89 & 76.04 & 83.61 & 76.08 & 84.74 & 74.72 & 82.75 & 82.81 & 83.31
      & +0.18 \\
      \textbf{+ Ours}
      & \textbf{56.19} & \textbf{95.22} & \textbf{55.95} & \textbf{93.00} & \textbf{61.63} & \textbf{93.97} & \textbf{84.95} & \textbf{76.30} & \textbf{83.68} & \textbf{76.92} & \textbf{85.33} & \textbf{75.40} & \textbf{83.62} & \textbf{83.74} & \textbf{84.23}
      & \textbf{+0.99} \\
      \bottomrule
    \end{tabular}
  \end{sc}
  }
\end{table*}

\section{Experiments}

We evaluate OATTA in two settings: a real-world evaluation on five diverse datasets spanning image, language, and wearable sensor modalities, including smartphone triaxial accelerometer data streams and single-channel EEG streams; and a controlled analysis on CIFAR-10~\citep{krizhevsky2009learning} variants to analyze behavior under known transition dynamics.

\subsection{Experimental Setup}

\textbf{Backbones and Training.}
To demonstrate the versatility of OATTA, we evaluate it in three distinct modalities. For all experiments, backbones were trained on the source training sets using standard cross-entropy loss. Specifically, the image models ResNet50~\citep{he2016deep}, ViT Base~\citep{dosovitskiy2020image}, and ConvNeXt Tiny~\citep{liu2022convnet}, as well as the language models TinyBERT (T-BERT)~\citep{jiao2020tinybert}, DistilBERT (D-BERT)~\citep{sanh2019distilbert}, and BERT Base (BERT)~\citep{devlin2019bert} were fine-tuned from pre-trained weights. In contrast, the (wearables) sensor models 1D CNN (CNN)~\citep{lecun1998convolutional}, TCN~\citep{lea2016temporal}, and 1D ResNet18~\citep{he2016deep} were trained from scratch on their respective source domains. Further details of the training hyperparameters are provided in Appendix~\ref{app:training}.

\textbf{Baselines.}
We evaluate OATTA both against and as a complementary module for three categories of baselines. First, the source (Base) uses standard inference with source-tuned weights. Second, we evaluate training-free ensembles: Monte Carlo Dropout (MC-Dr)~\citep{gal2016dropout} (averaging multiple stochastic forward passes) and Test-Time Augmentation (TTAug)~\citep{krizhevsky2012imagenet} (averaging augmented views). For architectures without dropout, we insert dropout layers during source training to enable MC-Dr at test time. Third, we evaluate gradient-based adaptation methods such as Tent~\citep{wang2020tent} and MEMO~\citep{zhang2022memo}. For Tent, we perform one gradient update per sample and reset the model to the source weights every 100 samples, enabling short-term adaptation while limiting long-term error accumulation. In contrast, we evaluate MEMO in an episodic protocol (reset per sample), minimizing marginal entropy over augmentations without carrying information across time. This setup separates online weight adaptation across the stream (Tent) from per-input augmentation consistency (MEMO).

\textbf{Wrapper Implementation.}
A key feature of OATTA is its modularity. In our experiments, we treat the baseline methods as prediction sources. For a baseline $B$ (e.g., Tent) producing a probability vector $q_t^{B}$, we first convert it to an observational likelihood $\mathcal{L}_t^B$ and then feed it into our recursive filter. This allows us to quantify the gain of temporal modeling over feature alignment.

\subsection{Datasets}

\subsubsection{Real-World Benchmarks}
For all real-world datasets, we adopt a user-centric split: we identify users or camera locations with the longest continuous data streams to serve as the test set, ensuring sufficient temporal depth for adaptation analysis. Detailed splitting protocols are provided in Appendix~\ref{app:dataset}.

\textbf{1. Camera Traps (image).}
We evaluate on the Caltech Camera Traps (CCT)~\citep{beery2018recognition} and UNSW Predators (UNSW)~\citep{unsw_predators_2025} datasets. Each stream corresponds to a single camera location. Dynamics reflect animal behavior (e.g., group motion and predator--prey sequences).

\textbf{2. Physiological Signals (sensor).}
We classify human states from smartphone triaxial accelerometer data using the Human Activity Recognition Trondheim (HARTH)~\citep{logacjov2021harth} dataset for activity recognition (e.g., walking, running) and from single-channel EEG streams using Sleep-EDF (Sleep)~\citep{kemp2000analysis} for sleep staging (e.g., N1, N2, N3). Each stream corresponds to a single human subject. These domains are governed by physical constraints, resulting in highly structured activity sequences rather than random state jumps.

\textbf{3. Social Media Sentiment (language).}
We perform binary sentiment classification on the Sentiment140 (SENT)~\citep{go2009twitter} dataset, which consists of textual tweets. Each stream corresponds to the chronological timeline of a single user. The dynamics reflect temporal inertia, e.g., a user posting negative content is more likely to continue posting negatively in the short term, though specific transition probabilities vary significantly across users.

\subsubsection{Controlled Evaluation}

We train a standard ResNet18 backbone on the original CIFAR-10 training set and utilize two test sets to simulate distinct adaptation scenarios.

\textbf{1. CIFAR-10.1 (v6)~\citep{recht2018cifar}.}
A reproduction of the CIFAR-10 test set containing exactly 2,000 new images sampled from the TinyImages dataset. This dataset was constructed to minimize distribution shift relative to the original CIFAR-10 while ensuring class balance, representing a realistic natural test scenario.

\textbf{2. CIFAR-10-C (corruptions)~\citep{hendrycks2019benchmarking}.}
A benchmark for evaluating classifier robustness against common perturbations. We employ the highest severity (level 5) corruptions, leveraging the varying difficulty of 15 corruption types as a mechanism to evaluate OATTA across the full spectrum of base model reliability, from severe failure (chance-level) to high accuracy. This allows us to empirically map the correlation between base model utility and OATTA adaptation gain.

\begin{table*}[t]
\centering
\caption{\textbf{Main results on controlled streams.} Accuracy (\%) is averaged over 10 runs; $\Delta$ Avg is mean$\pm$std over the same runs. Bold indicates the best performance in each base-vs-ours pair. Asterisks ($^*$) indicate statistically significant improvement ($p < 0.05$) by a two-sided Wilcoxon signed-rank test with Holm correction across the five baseline pairs (10 seeds). \textit{Stream definitions:} S1 (random) represents unstructured random sampling; S2 (sticky) simulates bursty temporal dynamics, where $\alpha$ denotes the self-transition probability (stickiness), higher $\alpha$ implies stronger consistency; S3 (permuted) simulates cyclic, workflow-like transitions; S4 (regime switch) and S5 (three-phase) introduce non-stationarity by shifting transition rules during the stream. OATTA robustly improves performance across diverse baselines and stream types, reaching 6.35\% gains on high-consistency streams, with negligible additional computational cost.}
\label{tab:accuracy_summary_extended}
\resizebox{\textwidth}{!}{
\begin{tabular}{l|cccccccccc|c}
\toprule
\textbf{Stream} & \textbf{Base} & \textbf{Ours} & \textbf{MC-Dr} & \textbf{+Ours} & \textbf{TTAug} & \textbf{+Ours} & \textbf{Tent} & \textbf{+Ours} & \textbf{MEMO} & \textbf{+Ours} & \textbf{$\Delta$ Avg} \\ \midrule
S1: random              & \textbf{77.05} & 76.81 & \textbf{76.99} & 76.81 & \textbf{78.70} & 78.59 & \textbf{77.00} & 76.85 & \textbf{79.11} & 79.02 & -0.15 $\pm$ 0.22\\
S2: sticky              & 76.98 & \sig{\textbf{78.56}} & 76.96 & \sig{\textbf{78.55}} & 78.83 & \sig{\textbf{81.50}} & 76.94 & \sig{\textbf{78.27}} & 79.20 & \sig{\textbf{79.94}} & 1.58 $\pm$ 0.23\\
S2: sticky ($\alpha=0.10$)    & \textbf{77.24} & 76.89 & \textbf{77.15} & 76.86 & \textbf{79.27} & 78.99 & \textbf{77.20} & 77.13 & \textbf{79.54} & 79.27 & -0.25 $\pm$ 0.21\\
S2: sticky ($\alpha=0.30$)    & \textbf{76.67} & 76.58 & \textbf{76.64} & 76.51 & 78.47 & \textbf{78.78} & \textbf{76.66} & 76.61 & \textbf{78.61} & 78.57 & -0.00 $\pm$ 0.23\\
S2: sticky ($\alpha=0.50$)    & 77.07 & \sig{\textbf{77.61}} & 77.03 & \sig{\textbf{77.61}} & 78.84 & \sig{\textbf{80.15}} & 77.07 & \sig{\textbf{77.53}} & 79.07 & \sig{\textbf{79.37}} & 0.64 $\pm$ 0.25\\
S2: sticky ($\alpha=0.70$)    & 76.98 & \sig{\textbf{78.56}} & 76.95 & \sig{\textbf{78.52}} & 79.05 & \sig{\textbf{81.69}} & 76.94 & \sig{\textbf{78.27}} & 79.23 & \sig{\textbf{80.00}} & 1.57 $\pm$ 0.30\\
S2: sticky ($\alpha=0.85$)    & 76.80 & \sig{\textbf{79.70}} & 76.71 & \sig{\textbf{79.72}} & 78.78 & \sig{\textbf{83.62}} & 76.78 & \sig{\textbf{79.10}} & 78.91 & \sig{\textbf{80.32}} & 2.90 $\pm$ 0.28\\
S2: sticky ($\alpha=0.90$)    & 76.92 & \sig{\textbf{80.62}} & 76.81 & \sig{\textbf{80.57}} & 79.29 & \sig{\textbf{85.56}} & 76.89 & \sig{\textbf{79.93}} & 79.31 & \sig{\textbf{81.17}} & 3.73 $\pm$ 0.56\\
S2: sticky ($\alpha=0.95$)    & 77.37 & \sig{\textbf{82.87}} & 77.28 & \sig{\textbf{82.92}} & 79.40 & \sig{\textbf{87.67}} & 77.36 & \sig{\textbf{82.03}} & 79.75 & \sig{\textbf{82.31}} & 5.33 $\pm$ 0.49\\
S2: sticky ($\alpha=0.98$)    & 77.05 & \sig{\textbf{83.48}} & 76.92 & \sig{\textbf{83.52}} & 78.67 & \sig{\textbf{88.70}} & 76.94 & \sig{\textbf{82.38}} & 79.02 & \sig{\textbf{82.24}} & 6.35 $\pm$ 0.37\\
S3: permuted            & 76.97 & \sig{\textbf{78.14}} & 76.94 & \sig{\textbf{78.12}} & 78.75 & \sig{\textbf{80.69}} & 76.98 & \sig{\textbf{77.84}} & 78.95 & \sig{\textbf{79.65}} & 1.17 $\pm$ 0.21\\
S4: regime switch       & 77.17 & \sig{\textbf{78.14}} & 77.09 & \sig{\textbf{78.11}} & 79.26 & \sig{\textbf{81.26}} & 77.15 & \sig{\textbf{77.94}} & 79.38 & \sig{\textbf{79.86}} & 1.05 $\pm$ 0.24\\
S5: three-phase         & 77.26 & \sig{\textbf{78.27}} & 77.05 & \sig{\textbf{78.29}} & 79.17 & \sig{\textbf{80.90}} & 77.19 & \sig{\textbf{78.00}} & 79.36 & \sig{\textbf{79.86}} & 1.06 $\pm$ 0.26\\
\bottomrule
\end{tabular}
}
\end{table*}

\textbf{3. Streaming Protocols (controlled generation).}
To isolate specific adaptation capabilities, we resample CIFAR-10 test sets into sequences with controlled dynamics:
\begin{itemize}[leftmargin=*, noitemsep, topsep=0pt]
    \item \textbf{S1: Random.} Samples are drawn independently with a uniform class distribution. This evaluates whether OATTA preserves statistical parity with the baseline in the absence of temporal order.

    \item \textbf{S2: Sticky.} Simulates bursty consistency where state $i$ tends to persist ($A_{ii}=\alpha$). The remaining probability mass is distributed uniformly across off-diagonal elements ($A_{ij} = \frac{1-\alpha}{K-1}$). Higher $\alpha$ implies stronger stickiness.
    
    \item \textbf{S3: Permuted.} Simulates workflow dynamics ($A~\to~B~\to~C$) by cyclically permuting the diagonal. We use a doubly stochastic transition matrix to ensure the stationary distribution remains uniform, isolating sequence learning from label shift.
    
    \item \textbf{S4 \& S5: Non-stationary.} \textit{Regime switch (S4)} abruptly changes the stickiness parameter of the S2 process at $T/2$. \textit{Three-phase (S5)} smoothly interpolates between two distinct S3 permutation rules, testing plasticity.
\end{itemize}

For each controlled stream setting, we average results over 10 runs with different random seeds controlling the stream permutation ordering.

\subsection{Results}
\subsubsection{Real-World Evaluation on Diverse Benchmarks}

We evaluate OATTA on five real-world datasets—CCT, UNSW, HARTH, Sleep, and SENT—with natural temporal ordering. The results are summarized in Table~\ref{combined_vertical_no_horiz}.

\textbf{Universality Across Modalities.}
OATTA demonstrates consistent improvements across all three modalities, improving the baseline model in 72 out of 75 cases. Whether applied to animal movement patterns in images, activity sequences in smartphone triaxial accelerometer and EEG streams, or sentiment consistency in language, the method effectively leverages the underlying order to improve accuracy. Notably, OATTA alone yields a mean gain of +1.49\% over Base—the largest among all compared methods, including the strongest baseline, TTAug (+1.31\%) (Table~\ref{combined_vertical_no_horiz}, $\Delta_{\text{Base}}$). Moreover, OATTA is complementary: when stacked on top of existing predictors, it delivers additional improvements and achieves the best overall result when combined with TTAug ($\Delta_{\text{Base}} = +2.39$).

\textbf{Effect of the LLR Gate.}
All results in Table~\ref{combined_vertical_no_horiz} are reported without the LLR. Appendix~\ref{app:llr_gate_eval_real} compares performance with and without the LLR: enabling it leads to a slight decrease on average on these naturally ordered benchmarks, since the gate conservatively interpolates toward the base predictor. When prior knowledge indicates meaningful temporal dependence in the stream, disabling the LLR is typically preferable to maximize gains.

\textbf{Signal Strength and Gain Variance.}
Consistent with the controlled results reported in Section~\ref{sec:controlled_results}, the magnitude of the real-world improvements correlates with the intrinsic ``stickiness'' (an implicit $\alpha$) of the data stream. We observe the largest gains for CCT (+1.84\% on Resnet50) and SENT (+1.38\% on T-BERT). This suggests these domains possess high inherent predictability: animals moving in groups and user mood inertia provide strong, high-$\alpha$ signals that OATTA readily exploits, while remaining lightweight and incurring negligible computational overhead.

\begin{figure}[t]
    \centering
    \includegraphics[width=0.85\linewidth]{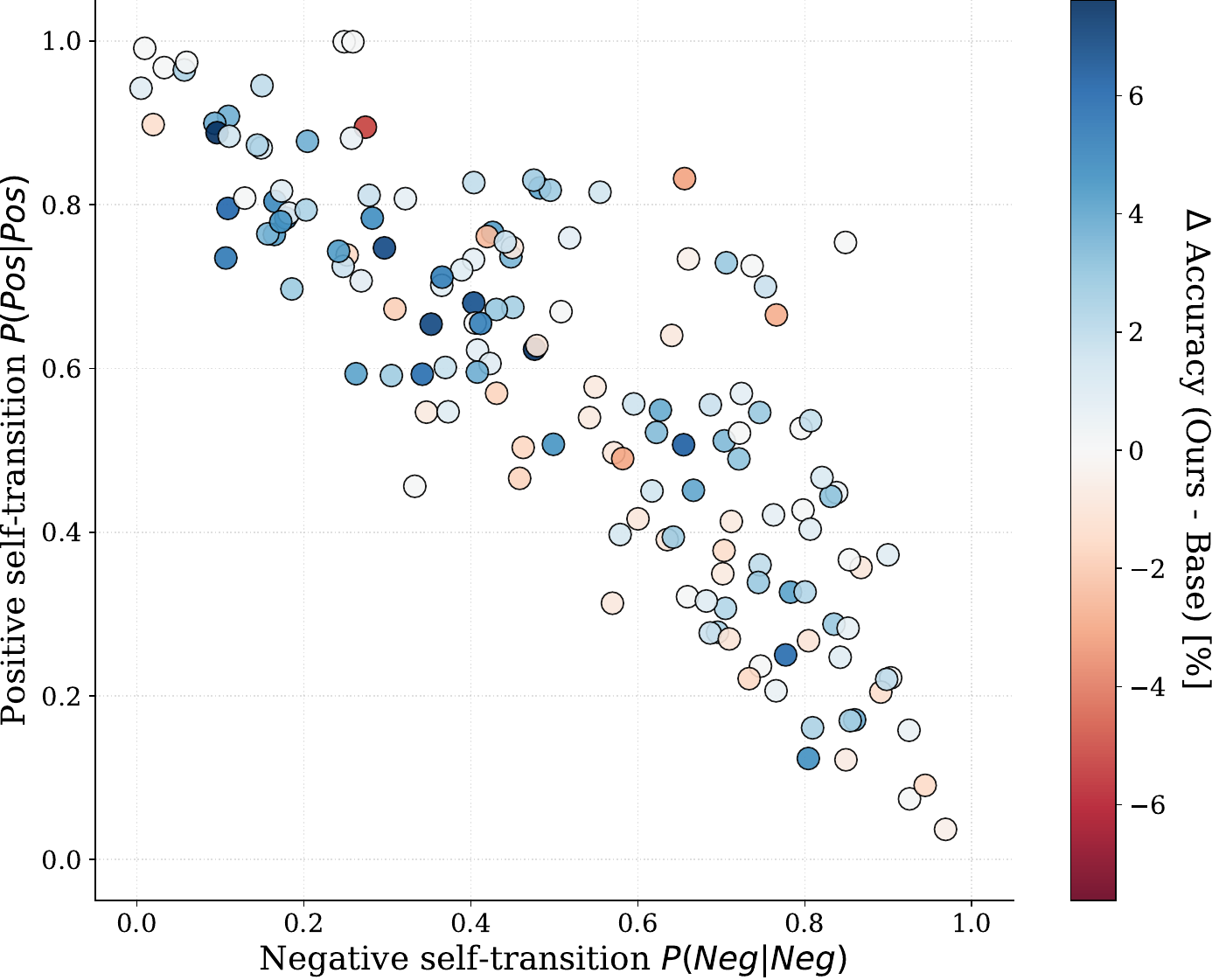} 
    \caption{\textbf{Impact of behavioral diversity on adaptation (SENT).} We visualize the learned transition dynamics ($p(Pos|Pos)$ vs. $p(Neg|Neg)$) for 150 unique users. The wide dispersion of points demonstrates significant heterogeneity, invalidating the utility of a single, universal transition rule. Color denotes the per-user accuracy change relative to the base model, $\Delta = \mathrm{Acc}(\text{Ours}) - \mathrm{Acc}(\text{Base})$ (blue: positive gain; red: negative gain).}
    \label{fig:user_diversity}
\end{figure}

\textbf{Behavioral Heterogeneity Analysis.}
To validate the need for user-specific adaptation, we examine the learned transition matrices for 150 users in the SENT dataset. As shown in Figure~\ref{fig:user_diversity}, the user dynamics exhibit substantial heterogeneity: users form distinct behavioral clusters, ranging from those with strong positive consistency to those with high negative consistency, with no single ``average'' profile that represents the population. This dispersion confirms that a static global prior would be suboptimal. Color encodes the per-user change in accuracy relative to the base model, $\Delta = \mathrm{Acc}(\text{Ours}) - \mathrm{Acc}(\text{Base})$; gains are mostly positive, and negative cases are scattered rather than localized, indicating benefits across transition tendencies rather than a single behavioral profile.

\textbf{Additivity to Baselines.}
A key finding is that OATTA provides additive gains when wrapped around various existing baselines, regardless of whether it uses gradient updates or simple heuristics. This can be attributed to OATTA providing an orthogonal source of information. Baselines such as Tent or TTAug focus on adapting model features or augmenting the input to handle distribution shift, whereas OATTA adapts the output probability sequence using temporal context. Consequently, combining feature-level adaptation with output-level temporal filtering consistently yields superior performance, enabling gains even on highly saturated baselines.

\subsubsection{Performance Comparison on Controlled Data Streams}
\label{sec:controlled_results}

Table~\ref{tab:accuracy_summary_extended} reports the classification accuracy and the average improvement ($\Delta$ Avg), defined as the percentage difference between the OATTA-adapted model and its corresponding baseline with the LLR. Appendix~\ref{app:llr_gate_eval_cont} compares the performance with and without the LLR in controlled data streams. Statistical significance is assessed with a two-sided Wilcoxon signed-rank test~\citep{wilcoxon1945individual} with Holm correction~\citep{holm1979simple} across the five baseline pairs per stream ($p<0.05$) (10 runs; Appendix~\ref{app:test}).

\textbf{Sensitivity to Temporal Strength.}
We observe an overall monotonic trend between the strength of the temporal signal (stickiness $\alpha$) and the adaptation gain, as detailed next.
\begin{itemize}[leftmargin=*, noitemsep]
    \item \textbf{Random Streams ($\alpha \to 0$).} On unstructured streams (S1) and low-correlation Markov streams ($\alpha \le 0.3$), OATTA yields a marginal numerical fluctuation ($< 0.25\%$). These differences are not statistically significant ($p>0.05$). This validates that the method maintains statistical parity with the baseline in the absence of temporal order, effectively converging to the marginal distribution without introducing noise.
    
    \item \textbf{Strong Signals ($\alpha \to 1$).} As $\alpha$ increases or when distinct latent structures are present (S3, S4, and S5), OATTA acts as a powerful performance multiplier. In these structured regimes (specifically S2 with $\alpha \ge 0.5$, S3, S4, and S5), statistical analysis confirms significant gains ($p<0.05$) across all structured settings. At $\alpha=0.98$, we achieve a peak gain of +6.35\%, substantially larger than the typical gain of +1\% or +2\% of standard TTA baselines in this setting—showing that, when a reliable signal exists, the transition prior $A_t$ effectively corrects base model errors.
\end{itemize}

Because the computational cost is negligible and OATTA remains statistically indistinguishable from the baseline when the test data exhibit limited to no temporal correlation, these results indicate that TTA should include transition-prior models such as the one introduced here by default.

\textbf{Sticky vs. Permuted Dynamics.}
Comparing the S2 (sticky) and S3 (permuted) streams reveals a limitation of count-based estimation. While OATTA improves S3 by +1.17\%, this is lower than the gains on comparable S2 streams. This performance gap stems from the recursive formulation of the EMA update. In sticky streams ($i \to i$), valid transitions occur consecutively, rapidly accumulating mass in the count matrix. In permuted streams ($i \to j$), valid transitions are interleaved; if the cycle period is long relative to the window size, the signal becomes diluted, making it harder for the count matrix to lock onto the deterministic trajectory.

\textbf{Robustness to Non-stationary Dynamics.}
We evaluate OATTA's plasticity under abrupt regime shifts using two protocols: S4 (regime switch) and S5 (three-phase). Despite the lag inherent in EMA, OATTA maintains positive gains in both scenarios (+1.05\% and +1.06\%, respectively).

\begin{figure}[t]
    \centering
    \includegraphics[width=0.85\linewidth]{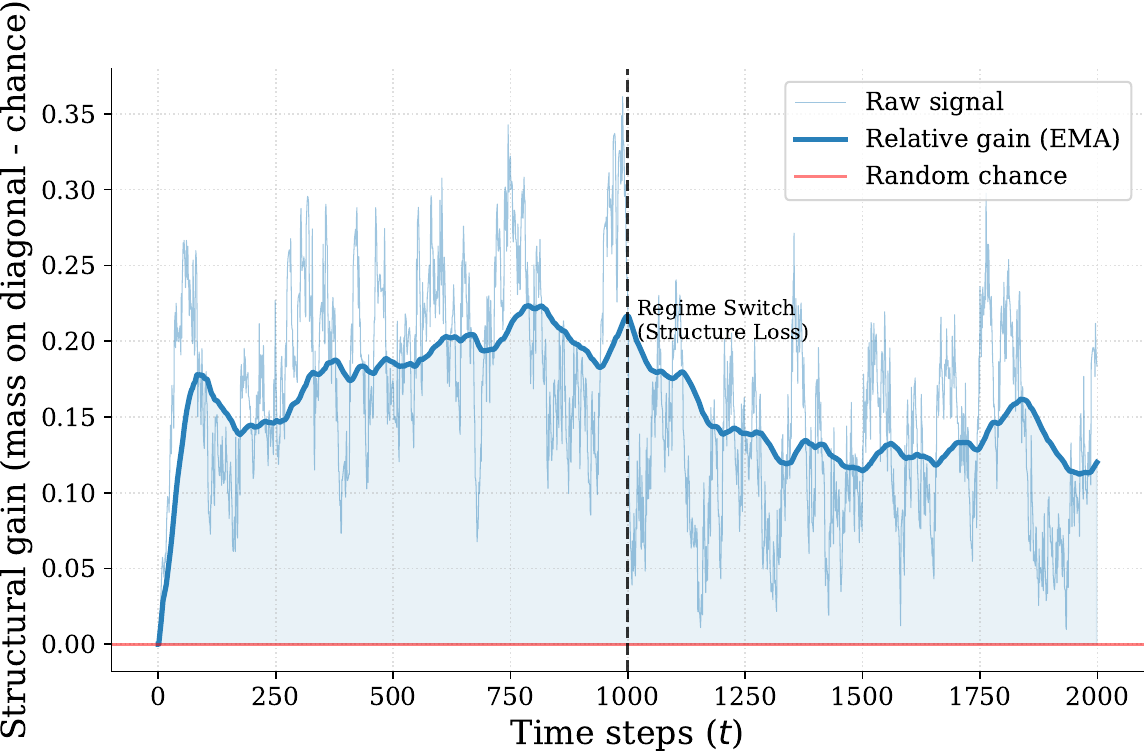}
    \caption{\textbf{Adaptation plasticity under regime shift ($\alpha=0.7 \to 0.5$).} We track the structural gain on the CIFAR-10 stream. The light blue trace shows the noisy base signal, while the solid line (EMA, span=300) reveals the adaptation trend. At $t=1{,}000$, the stream becomes less predictable; OATTA detects this shift and rapidly down-calibrates its prior strength, stabilizing at a new equilibrium that accurately reflects the reduced temporal signal.}
    \label{fig:plasticity}
\end{figure}

To investigate the mechanism behind this recovery, we visualize the adaptation to a regime switch ($\alpha=0.7 \to 0.5$) in Figure~\ref{fig:plasticity}. We track the structural gain $G_t$, defined as the diagonal probability mass of the learned transition matrix in excess of random chance ($G_t = \frac{1}{K}\sum_{i=1}^K A_{ii} - \frac{1}{K}$). Initially, the system locks onto the strong signal ($\alpha=0.7$), maintaining a high gain ($\approx +0.2$). Following the shift at $t=1{,}000$, OATTA acts as a calibrated estimator: rather than collapsing or remaining over-confident, it immediately down-calibrates its prior strength, stabilizing at a new equilibrium ($\approx +0.15$) that accurately reflects the reduced temporal signal.

\textbf{LLR Impact.}
Appendix~\ref{app:llr_gate_eval_cont} reports a controlled-stream comparison of OATTA \emph{with} and \emph{without} the LLR. The LLR improves robustness in weakly structured or order-agnostic streams: on S1/random, the average change is close to zero with the LLR ($\Delta_{\text{Avg}}=-0.15\%$) but decreases substantially without the LLR ($\Delta_{\text{Avg}}=-3.92\%$). In contrast, when temporal dependence is strong, disabling the LLR yields larger gains (e.g., at $\alpha=0.98$, $\Delta_{\text{Avg}}=12.23\%$ without the LLR vs.\ $6.35\%$ with the LLR), since the filter can fully exploit the stream's predictive dynamics. Accordingly, we use the ungated variant in our main controlled experiments and provide a comparative analysis with and without LLR in the appendix for completeness.



\subsubsection{Sensitivity Analysis: The Adaptation Threshold}

To understand the boundary conditions of our method, we investigate the dependence of the adaptation gain on the quality of the base model. We utilize the CIFAR-10-C dataset (level 5), which consists of 15 distinct corruption types (e.g., Gaussian noise, motion blur). Since each corruption degrades the base model to a different extent, this dataset allows us to sweep a wide range of base accuracies.

\begin{figure}[t]
    \centering
    \includegraphics[width=0.85\linewidth]{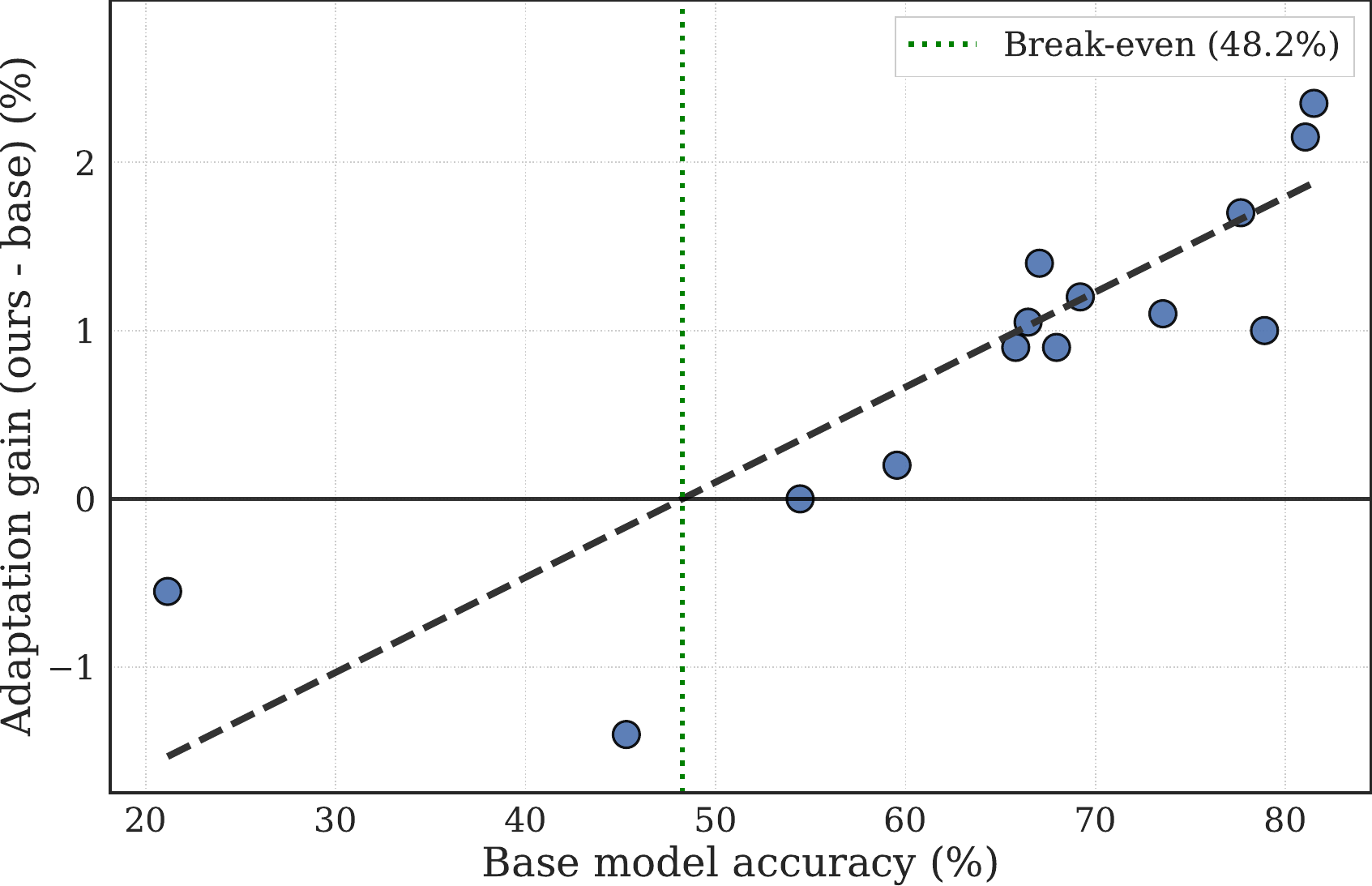}
    \caption{\textbf{Base model utility vs. adaptation gain.} Correlation analysis on CIFAR-10-C (level 5) evaluated on S2 (sticky) streams. Each point represents one of the 15 corruption types. We observe a strong linear correlation ($R \approx 0.82$), with an empirical break-even point at 48.2\% accuracy.}
    \label{fig:correlation}
\end{figure}

As shown in Figure~\ref{fig:correlation}, we observe a strong linear correlation ($R \approx 0.82$) between base accuracy and adaptation gain, which is precisely fitted by the linear model.
Solving for $\text{Gain}=0$ identifies an empirical break-even threshold at 48.2\% accuracy. This boundary reflects the minimum signal-to-noise ratio required for effective self-supervision: below 48.2\%, the transition matrix aggregates noise from wrong predictions, thereby degrading performance. Above this level, the model provides sufficient ``anchor'' predictions for OATTA to estimate valid dynamics, acting as a consistent performance multiplier.

\section{Conclusion}

In this work, we identified a critical blind spot in current TTA research, namely the treatment of streaming data as unordered collections of samples, ignoring the rich supervisory signal provided by temporal dynamics. To address this, we proposed OATTA, a lightweight model-agnostic framework that formulates adaptation as a recursive Bayesian estimation problem. We additionally propose an optional LLR that quantifies when the temporal structure is unsupported and triggers a conservative fallback toward the base predictor. In practice, omitting LLR is preferable when the deployment stream is expected to exhibit a meaningful temporal dependence, whereas enabling LLR is a more reliable default under uncertainty, as it avoids statistically significant degradation on near-i.i.d.\ streams while still capturing gains when the temporal signal is present. 

Our extensive empirical evaluation demonstrates that OATTA is both universal and additive, exploiting a source of information that standard baselines ignore. Across diverse modalities—ranging from animal behavior in camera traps and human activity in wearable sensors to sentiment consistency in social media—our method yields robust performance gains, validating that temporal correlations are a domain-agnostic resource. Crucially, OATTA operates as an orthogonal performance multiplier; it does not replace feature-adaptation methods like Tent or MEMO but complements them. By combining feature-level alignment with output-level temporal filtering, we achieve superior results.



\section*{Acknowledgments} 
Work partially supported by NSF, ONR, and the Simons Foundation. 

\section*{Impact Statement}

This paper proposes Order-Aware Test-Time Adaptation (OATTA), a lightweight, model-agnostic module that improves robustness by leveraging temporal dependencies within incoming streams. The intended impact is positive: OATTA enables more reliable deployment of machine learning systems in continuously operating environments (e.g., wearable and physiological monitoring, long-running vision systems such as camera traps, and sequential text classification) with minimal additional computational overhead and without requiring access to source training data at test time. By improving stability and accuracy under input uncertainty, OATTA can reduce the need for frequent retraining or manual relabeling, lowering deployment costs and energy consumption.

Potential risks include misuse and failure modes. Improved streaming inference could enable socially harmful applications (e.g., surveillance or behavioral profiling) if deployed without appropriate governance. Because OATTA learns stream-specific transition dynamics, the learned transition matrix may implicitly encode behavioral routines (e.g., daily activities or persistent sentiment patterns); if it is logged or transmitted, it can raise privacy and security concerns even when raw inputs are not stored. Finally, temporal priors can reinforce systematic errors: biases in the base predictor may be amplified over time, and when temporal structure is weak or corrupted, test-time adaptation may overfit spurious transitions and degrade performance.


\bibliography{example_paper}
\bibliographystyle{icml2026}

\newpage
\appendix
\onecolumn

\section{Prior Sensitivity and Label Shift}
\label{app:lds_robustness}

\textbf{Experimental Setup.} We analyze the impact of the prior assumption $\rho$ used in likelihood conversion ($\mathcal{L}_t \propto q_t / \rho$). We train a ResNet18 on an imbalanced CIFAR-10 where five classes are subsampled to 10\%. We compare two strategies: utilizing the source prior ($\rho = P_{src}$), effectively performing logit adjustment, versus our default uniform prior ($\rho = \mathcal{U}$), which assumes no metadata. We evaluate on two conditions: (i) \emph{label distribution shift}, where the target stream becomes approximately uniform over classes, and (ii) \emph{no shift}, where the target stream preserves the source imbalance from the imbalanced CIFAR-10 training set.

\paragraph{Stream generation.}
For each stream, labels are sampled sequentially from a first-order Markov chain over CIFAR-10 classes $\{C0,\ldots,C9\}$. Given the current label $y_t$, we sample the next label $y_{t+1}$ according to the transition probabilities in row $y_t$ of $A$:
\[
\Pr(y_{t+1}=j \mid y_t=i) = A_{ij},
\]
where $A$ is a row-stochastic transition matrix (each row sums to 1). We use two stream types in our analysis:
(i) S2 (sticky), which uses a single stationary transition matrix; and
(ii) S4 (regime switch), which uses one transition matrix for the first half of the stream and another transition matrix for the second half.

\paragraph{Label distribution shift condition.}
To induce label distribution shift, we use the sticky stream S2 with stickiness $\alpha=0.7$. For the regime-switch stream S4, we set $\alpha=0.7$ for the first half of the stream and reduce the temporal strength to $\alpha=0.5$ for the second half. This creates a change in the temporal dynamics that accompanies the shift toward a more uniform target stream.

\paragraph{No-shift condition (preserving source imbalance).}
To construct streams that maintain the source imbalance, we use structured transition matrices that keep the self-transition probability high while allocating more transition mass among the majority classes (class 0 to class 4) than the minority classes (class 5 to class 9). This design keeps the temporal strength comparable across S2 and S4 while preserving the intended class imbalance.

\paragraph{Transition matrices.}
For S2 (sticky, no shift), we use the following transition matrix:
\begin{center}
\small
\begin{tabular}{c|cccccccccc}
\hline
From$\backslash$To & Class 0 & Class 1 & Class 2 & Class 3 & Class 4 & Class 5 & Class 6 & Class 7 & Class 8 & Class 9 \\
\hline
Class 0 & .755 & .055 & .055 & .055 & .055 & .005 & .005 & .005 & .005 & .005 \\
Class 1 & .055 & .755 & .055 & .055 & .055 & .005 & .005 & .005 & .005 & .005 \\
Class 2 & .055 & .055 & .755 & .055 & .055 & .005 & .005 & .005 & .005 & .005 \\
Class 3 & .055 & .055 & .055 & .755 & .055 & .005 & .005 & .005 & .005 & .005 \\
Class 4 & .055 & .055 & .055 & .055 & .755 & .005 & .005 & .005 & .005 & .005 \\
Class 5 & .055 & .055 & .055 & .055 & .055 & .705 & .005 & .005 & .005 & .005 \\
Class 6 & .055 & .055 & .055 & .055 & .055 & .005 & .705 & .005 & .005 & .005 \\
Class 7 & .055 & .055 & .055 & .055 & .055 & .005 & .005 & .705 & .005 & .005 \\
Class 8 & .055 & .055 & .055 & .055 & .055 & .005 & .005 & .005 & .705 & .005 \\
Class 9 & .055 & .055 & .055 & .055 & .055 & .005 & .005 & .005 & .005 & .705 \\
\hline
\end{tabular}
\end{center}

For S4 (regime switch, no shift), we use the above matrix for the first half of the stream and the following matrix for the second half:
\begin{center}
\small
\begin{tabular}{c|cccccccccc}
\hline
From$\backslash$To & Class 0 & Class 1 & Class 2 & Class 3 & Class 4 & Class 5 & Class 6 & Class 7 & Class 8 & Class 9 \\
\hline
Class 0 & .5909 & .0909 & .0909 & .0909 & .0909 & .0091 & .0091 & .0091 & .0091 & .0091 \\
Class 1 & .0909 & .5909 & .0909 & .0909 & .0909 & .0091 & .0091 & .0091 & .0091 & .0091 \\
Class 2 & .0909 & .0909 & .5909 & .0909 & .0909 & .0091 & .0091 & .0091 & .0091 & .0091 \\
Class 3 & .0909 & .0909 & .0909 & .5909 & .0909 & .0091 & .0091 & .0091 & .0091 & .0091 \\
Class 4 & .0909 & .0909 & .0909 & .0909 & .5909 & .0091 & .0091 & .0091 & .0091 & .0091 \\
Class 5 & .0909 & .0909 & .0909 & .0909 & .0909 & .5091 & .0091 & .0091 & .0091 & .0091 \\
Class 6 & .0909 & .0909 & .0909 & .0909 & .0909 & .0091 & .5091 & .0091 & .0091 & .0091 \\
Class 7 & .0909 & .0909 & .0909 & .0909 & .0909 & .0091 & .0091 & .5091 & .0091 & .0091 \\
Class 8 & .0909 & .0909 & .0909 & .0909 & .0909 & .0091 & .0091 & .0091 & .5091 & .0091 \\
Class 9 & .0909 & .0909 & .0909 & .0909 & .0909 & .0091 & .0091 & .0091 & .0091 & .5091 \\
\hline
\end{tabular}
\end{center}

\begin{table}[t]
\centering
\caption{\textbf{Prior sensitivity analysis.} Comparison of test accuracy (\%) using source prior ($P_{src}$) or uniform prior ($\mathcal{U}$). The source prior excels when the label distribution shifts, while the uniform prior preserves performance when the stream remains imbalanced.}
\label{tab:prior_robustness}
\begin{tabular}{l|cccc}
\toprule
& \multicolumn{2}{c}{Label distribution shift} & \multicolumn{2}{c}{No shift} \\
\cmidrule(lr){2-3} \cmidrule(lr){4-5}
\textbf{Prior choice} & S2 & S4 & S2 & S4 \\
\midrule
Base model & 62.30 & 63.15 & 73.55 & 73.50 \\
\textit{Source prior ($P_{src}$)} & \textbf{64.85} & \textbf{64.90} & 74.50 & 74.00 \\
\midrule
\textit{Uniform prior ($\mathcal{U}$)} & 63.50 & 64.15 & \textbf{75.25} & \textbf{74.45} \\
\bottomrule
\end{tabular}

\end{table}

\textbf{Results \& Discussion.} Table~\ref{tab:prior_robustness} reveals a clear trade-off dictated by the test environment. Under label distribution shift, using the source prior ($\rho = P_{src}$) consistently outperforms the uniform prior by effectively neutralizing the source bias. Conversely, under no shift, the uniform prior proves superior, as dividing by $P_{src}$ incurs an unnecessary penalty by suppressing the valid majority class bias. Crucially, however, even under label distribution shift, the uniform prior still consistently outperforms the base model, validating its ability to provide adaptation gains without source statistics.

This suggests a strategic distinction: if source metadata is available and a shift is known, setting $\rho = P_{src}$ is effective. However, in ``black-box'' or ``source-free'' scenarios where metadata is unavailable or the target distribution is unknown, the uniform prior ($\rho = \mathcal{U}$) serves as a robust default.

\section{Likelihood-Ratio Gate: Implementation Details}
\label{app:llr_gate}

This appendix provides the full specification of the likelihood-ratio gate, introduced in Section~\ref{sec:llr_gate}, including the order-agnostic baseline prior, efficient evidence accumulation, and complete pseudocode.

\subsection{Order-Agnostic Baseline Prior}
\label{app:llr_baseline_prior}
The gate compares the temporal prior $\pi_t$ to an order-agnostic baseline prior $\bar{\pi}_t$ that does not use temporal transitions. In our implementation, $\bar{\pi}_t$ is a smoothed estimate of class frequencies computed via an exponential moving average of model predictions:
\begin{equation}
\bar{\pi}_t \;=\; (1-\eta)\,\bar{\pi}_{t-1} \;+\; \eta\, q_t,
\qquad \bar{\pi}_t \leftarrow \frac{\bar{\pi}_t}{\sum_{k=1}^K \bar{\pi}_t(k)},
\end{equation}
where $\eta\in(0,1]$ controls the adaptation speed of the baseline (larger $\eta$ tracks the current stream more aggressively). We initialize $\bar{\pi}_0=\mathbf{1}/K$.

\subsection{Evidence Accumulation}
\label{app:llr_window}
Section~\ref{sec:llr_gate} defines a windowed average of the per-step evidence $\Delta_t$. In practice, we compute an equivalent exponentially weighted estimator that avoids storing a length-$W$ buffer:
\begin{equation}
\mathrm{LLR}_t \;=\; \left(1-\frac{1}{W}\right)\mathrm{LLR}_{t-1} \;+\; \frac{1}{W}\,\Delta_t,
\qquad \mathrm{LLR}_0 = 0,
\end{equation}
where $W$ sets the effective memory length (larger $W$ produces a more stable but slower-responding gate). The per-step evidence is
\begin{equation}
\Delta_t
=\log\!\big(\langle q_t,\pi_t\rangle+\varepsilon\big)
-\log\!\big(\langle q_t,\bar{\pi}_t\rangle+\varepsilon\big),
\end{equation}
with $\varepsilon>0$ for numerical stability.

\subsection{Complete Gated Algorithm}
\label{app:llr_pseudocode}
Algorithm~\ref{alg:oatta_llr} lists the full procedure, including the gate and the baseline-prior update.

\begin{algorithm}[t]
\caption{OATTA with Likelihood-Ratio Gate}
\label{alg:oatta_llr}
\begin{algorithmic}[1]
\STATE {\bfseries Input:} stream $\mathcal{S}$, predictor $f_\theta$, class prior $\rho$ (default: uniform)
\STATE {\bfseries Init:} transition counts $C_0 \leftarrow \kappa I$, $A_0 \leftarrow \mathrm{RowNorm}(C_0)$
\STATE \hspace{1.35em} filtered posterior $p_0 \leftarrow \mathbf{1}/K$, baseline prior $\bar{\pi}_0 \leftarrow \mathbf{1}/K$, evidence $\mathrm{LLR}_0 \leftarrow 0$
\FOR{$t=1$ {\bfseries to} $T$}
    \STATE $q_t \leftarrow f_\theta(x_t)$
    \STATE $\pi_t \leftarrow A_{t-1}^\top p_{t-1}$
    \STATE $p_t \propto (q_t/\rho)\odot \pi_t$ \;\; (normalize)

    \STATE $\bar{\pi}_t \leftarrow (1-\eta)\bar{\pi}_{t-1} + \eta q_t$ \;\; (normalize)
    \STATE $\Delta_t \leftarrow \log(\langle q_t,\pi_t\rangle+\varepsilon) - \log(\langle q_t,\bar{\pi}_t\rangle+\varepsilon)$
    \STATE $\mathrm{LLR}_t \leftarrow (1-\frac{1}{W})\mathrm{LLR}_{t-1} + \frac{1}{W}\Delta_t$
    \STATE $\lambda_t \leftarrow \sigma\!\left(\frac{\mathrm{LLR}_t - m}{\tau}\right)$
    \STATE $\hat{p}_t \leftarrow \lambda_t p_t + (1-\lambda_t) q_t$ \;\; (normalize)

    \STATE $C_t \leftarrow (1 \!-\! \gamma w_t)C_{t-1} + \gamma w_t (q_{t-1} \otimes q_t)$ 
   \STATE $A_t \leftarrow \text{RowNorm}(C_t)$; \quad $p_{t-1} \leftarrow p_t; \quad q_{t-1} \leftarrow q_t$
    \STATE $p_t \leftarrow \hat{p}_t$
\ENDFOR
\end{algorithmic}
\end{algorithm}

\subsection{Hyperparameter Effects}
\label{app:llr_hparams}
The margin $m$ sets how much stronger the temporal explanation must be before the method trusts it; increasing $m$ makes the gate more conservative and improves robustness on nearly i.i.d.\ streams. The temperature $t$ controls the sharpness of the transition between relying on the Bayes-filtered posterior and reverting to the base predictor; smaller $t$ yields a more abrupt switch, while larger $t$ yields smoother interpolation. The window length $W$ controls the stability--reactivity trade-off of the evidence statistic.

\section{Complexity Computation}
\label{app:complexity_proof}

We quantify the computational complexity of the proposed method by analyzing the floating-point operations (FLOPs) required per time step $t$. Let $K$ be the number of classes. The OATTA algorithm~\ref{alg:oatta} consists of the following vector-space operations:

\begin{enumerate}
    \item \textbf{Prior Prediction ($\pi_t = A_{t-1}^\top p_{t-1}$):} Matrix--vector multiplication ($K \times K$ by $K \times 1$).
    \begin{itemize}
        \item FLOPs: $K^2$ multiplications + $K(K-1)$ additions $\approx 2K^2$.
    \end{itemize}
    
    \item \textbf{Measurement Update:} 
    \begin{itemize}
        \item Likelihood conversion ($\mathcal{L}_t \leftarrow q_t / \rho$): $K$ divisions.
        \item Fusion ($p_t \propto \mathcal{L}_t \odot \pi_t$): $K$ multiplications.
        \item Normalization ($p_t / \sum p_t$): $K$ additions (summation) + $K$ divisions.
        \item Subtotal: $\approx 4K$.
    \end{itemize}
    
    \item \textbf{Transition Estimation ($C_t$):}
    \begin{itemize}
        \item Entropy calculation ($H(q_t)$): $\approx 3K$ (logarithm, multiplication, summation).
        \item Outer product ($q_{t-1} \otimes q_t$): $K^2$ multiplications.
        \item EMA update ($C_t \leftarrow (1-\gamma w_t)C_{t-1} + \dots$): $2K^2$ (scalar multiplication and matrix addition).
        \item Row normalization: $\approx 2K^2$ (row summation and division).
    \end{itemize}
    
    \item \textbf{Total OATTA FLOPs:} 
    $$ \text{Computational cost} \approx 2K^2 \text{ (Prior)} + 5K^2 \text{ (Transition)} + \mathcal{O}(K) \approx 7K^2. $$
\end{enumerate}

\noindent\textbf{Comparison Against the Backbone (ResNet50).}
A standard ResNet50 forward pass requires approximately 3.8 GFLOPs~\citep{he2016deep} ($3.8 \times 10^9$).

\begin{itemize}
    \item \textbf{UNSW ($K=5$):}
    $$ \text{Cost} \approx 7(5)^2 = 175 \text{ FLOPs} $$
    $$ \text{Overhead} = \frac{175}{3.8 \times 10^9} \approx 4.61 \times 10^{-8} $$

    \item \textbf{CIFAR-10 ($K=10$):}
    $$ \text{Cost} \approx 7(10)^2 = 700 \text{ FLOPs} $$
    $$ \text{Overhead} = \frac{700}{3.8 \times 10^9} \approx 1.84 \times 10^{-7} $$

    \item \textbf{CCT ($K=14$):}
    $$ \text{Cost} \approx 7(14)^2 = 1{,}372 \text{ FLOPs} $$
    $$ \text{Overhead} = \frac{1{,}372}{3.8 \times 10^9} \approx 3.61 \times 10^{-7} $$

    \item \textbf{ImageNet ($K=1000$):}
    $$ \text{Cost} \approx 7(1000)^2 = 7{,}000{,}000 \text{ FLOPs} $$
    $$ \text{Overhead} = \frac{7{,}000{,}000}{3.8 \times 10^9} \approx 1.84 \times 10^{-3} $$
\end{itemize}

In all cases, the computational cost of OATTA is orders of magnitude lower than the backbone inference cost, rendering the overhead negligible.

\section{Experimental Details}

\subsection{Training Details}
\label{app:training}

In this section, we detail the source training protocols for all backbones used in our experiments. 

\subsubsection{Controlled Evaluation (CIFAR-10)}
For the controlled analysis on CIFAR-10, we trained a ResNet18 architecture from scratch to avoid inductive bias from pretraining. To enable ensemble-based baselines (e.g., MC-Dropout) at test time, we modified the standard ResNet18 architecture by replacing the final fully connected layer with a sequence of \texttt{Dropout($p=0.2$) $\to$ Linear}.

We used a fixed hyperparameter configuration without grid search. The model was trained using stochastic gradient descent (SGD) for 100 epochs with the following parameters:
\begin{itemize}[noitemsep]
    \item \textbf{Batch size:} 128
    \item \textbf{Optimizer:} SGD with Nesterov momentum ($\mu=0.9$)
    \item \textbf{Learning rate:} Initialized at $0.1$, decayed by a factor of $0.1$ at epochs 50 and 75
    \item \textbf{Weight decay:} $5 \times 10^{-4}$
    \item \textbf{Data augmentation:} Standard random cropping ($32 \times 32$, padding 4) and horizontal flipping
\end{itemize}

\subsubsection{Real-World Benchmarks}
For the real-world benchmarks, the training strategy varied by modality.

\textbf{Image and Language (fine-tuning).} For image datasets (CCT, UNSW) and language datasets (SENT), we initialized the backbones using standard pre-trained weights. We fine-tuned all parameters using the AdamW optimizer.

\textbf{Sensor (training from scratch).} For sensor-based domains (HARTH, Sleep), where large-scale pre-trained models are not available, we trained the CNN, TCN, and 1D ResNet18 backbones from scratch.

\textbf{Hyperparameter selection.} Unlike CIFAR-10, we performed a grid search for all real-world datasets to select the optimal learning rate. We trained three learning rates per dataset (specific to the modality) and selected the checkpoint with the highest accuracy on a held-out validation set. The search spaces and final configurations are summarized in Table~\ref{tab:training_params}.

\begin{table*}[h]
    \centering
    \caption{\textbf{Source training hyperparameters.} We report the search space for learning rates (LR) and the training duration (Max steps) for each dataset. The best LR was selected via validation accuracy.}
    \label{tab:training_params}
    \begin{small}
    \begin{sc}
    \resizebox{0.9\textwidth}{!}{
    \begin{tabular}{l| l l l c c}
    \toprule
    \textbf{Dataset} & \textbf{Modality} & \textbf{Backbones} & \textbf{LR grid search} & \textbf{Steps} & \textbf{Weight decay} \\
    \midrule
    CCT & Image & ResNet50, ViT, ConvNeXt & $\{1\text{e-}4, 5\text{e-}5, 1\text{e-}5\}$ & 15{,}000 & $1\text{e-}4$ \\
    UNSW & Image & ResNet50, ViT, ConvNeXt & $\{1\text{e-}4, 5\text{e-}5, 1\text{e-}5\}$ & 5{,}000 & $1\text{e-}4$ \\
    HARTH & Sensor & CNN, TCN, 1D ResNet18 & $\{1\text{e-}3, 5\text{e-}4, 1\text{e-}4\}$ & 45{,}000 & $1\text{e-}3$ \\
    Sleep & Sensor & CNN, TCN, 1D ResNet18 & $\{1\text{e-}3, 5\text{e-}4, 1\text{e-}4\}$ & 45{,}000 & $1\text{e-}3$ \\
    SENT & Language & T-BERT, D-BERT, BERT & $\{1\text{e-}5, 2\text{e-}5, 3\text{e-}5\}$ & 30{,}000 & $1\text{e-}2$ \\
    \bottomrule
    \end{tabular}
    }
    \end{sc}
    \end{small}
\end{table*}

\subsection{Dataset Details}
\label{app:dataset}

\paragraph{Train/Validation/Test split protocol.}
Each dataset consists of temporal \emph{streams} that are naturally grouped by an \emph{identity} dimension: (i) \emph{camera location} for vision-based datasets (CCT, UNSW) and (ii) \emph{user/subject} for wearable, user-centric datasets (HARTH, Sleep, SENT). To prevent identity leakage, we construct the splits in two stages.

\begin{enumerate}
    \item \textbf{Top-$N$ longest streams as the test split.}
    From the full set of identities (users or camera locations), we rank identities by stream length and select the top-$N$ identities with the longest continuous streams as the test split. We exclude all samples from these test identities from the remaining data, ensuring that test identities never appear in training or validation.

    \item \textbf{Training/validation split (90/10).}
    We randomly split the remaining samples into 90\% training and 10\% validation at the sample level (i.e., identities may appear in both training and validation).
\end{enumerate}

\paragraph{Dataset-wise split statistics.}
Table~\ref{tab:dataset_split_summary} summarizes the number of identities and the resulting number of samples in the training, validation, and test splits.

\begin{table}[h]
    \centering
    \caption{Dataset split summary. ``Identity'' refers to camera locations (CCT, UNSW) or users/subjects (HARTH, Sleep, SENT). ``ID'' denotes the number of unique identities, and $N$ denotes the number of instances (samples) in each split.}
    \label{tab:dataset_split_summary}
    \begin{tabular}{l|l r r r r r r}
        \hline
        Dataset & Identity & ID (All) & ID (Test) & ID (Train+Validation) & $N$ (Train) & $N$ (Validation) & $N$ (Test) \\
        \hline
        CCT   & Camera location & 65  & 15  & 50      & 18{,}328   & 2{,}037   & 154{,}749 \\
        UNSW  & Camera location & 82  & 15  & 67      & 5{,}055    & 562       & 3{,}318 \\
        HARTH & User/subject    & 23  & 5   & 18      & 74{,}504   & 14{,}713  & 39{,}997 \\
        Sleep & User/subject    & 76  & 10  & 66      & 147{,}802  & 20{,}777  & 26{,}900 \\
        SENT  & User/subject    & 618{,}172 & 150 & 618{,}022 & 1{,}278{,}053 & 142{,}008 & 22{,}505 \\
        \hline
    \end{tabular}
\end{table}

\paragraph{HARTH: label shrinking (activity grouping).}
To reduce class fragmentation, we map the original HARTH activity labels into six broader groups (Walking, Running, Static, Lying, Stairs, Cycling). 

\paragraph{SENT: handling the rare neutral label.}
For SENT, we remove the neutral label because it has very few samples relative to the other classes, which would otherwise introduce severe class imbalance.

\section{Real-World Evaluation with and without LLR}
\label{app:llr_gate_eval_real}
\begin{table*}[t]
  \centering
  \caption{\textbf{Domain-centric view of universality.} We organize results by domain to highlight robustness across architectures and modalities. Bold \emph{numbers} indicate the best result within each block (baseline, without LLR, and with LLR). $\Delta_{\text{Base}}$ denotes the mean improvement relative to Base.}
  \label{combined_vertical_no_horiz2}
  \setlength{\tabcolsep}{1.8pt}
  \resizebox{\textwidth}{!}{%
  \begin{sc}
    \begin{tabular}{l| cc cc cc cc cc cc ccc | c}
      \toprule
      & \multicolumn{6}{c}{\textbf{Image Domain}}
      & \multicolumn{6}{c}{\textbf{Sensor Domain}}
      & \multicolumn{3}{c}{\textbf{Language Domain}}
      & \multicolumn{1}{c}{} \\
      \cmidrule(lr){2-7} \cmidrule(lr){8-13} \cmidrule(lr){14-16} 
      & \multicolumn{2}{c}{ResNet50} & \multicolumn{2}{c}{ViT Base} & \multicolumn{2}{c}{ConvNeXt}
      & \multicolumn{2}{c}{CNN} & \multicolumn{2}{c}{TCN} & \multicolumn{2}{c}{1D ResNet18}
      & T-BERT & D-BERT & BERT
      & $\Delta_{\text{Base}}$ \\
      \cmidrule(lr){2-3} \cmidrule(lr){4-5} \cmidrule(lr){6-7}
      \cmidrule(lr){8-9} \cmidrule(lr){10-11} \cmidrule(lr){12-13}
      \cmidrule(lr){14-14} \cmidrule(lr){15-15} \cmidrule(lr){16-16}
      Method
      & CCT & UNSW & CCT & UNSW & CCT & UNSW
      & HARTH & Sleep & HARTH & Sleep & HARTH & Sleep
      & SENT & SENT & SENT
      &  \\
      \midrule

      Base
      & 54.30 & 92.24 & 53.17 & 91.70 & 58.88 & 92.33 & 83.82 & 75.97 & 82.78 & 76.16 & 84.58 & 74.05 & 84.25 & 85.25 & 85.84
      & -- \\
      \textbf{without LLR}
      & 57.08 & 93.85 & 55.24 & \textbf{93.07} & 61.84 & \textbf{93.53} & \textbf{84.97} & 76.05 & \textbf{83.64} & 77.07 & \textbf{85.37} & \textbf{75.35} & \textbf{86.27} & \textbf{86.93} & \textbf{87.35}
      & \textbf{+1.49} \\
      \textbf{with LLR}
      & \textbf{57.26} & \textbf{93.95} & \textbf{55.30} & 93.04 & \textbf{61.89} & \textbf{93.53} & 84.69 & \textbf{76.53} & 83.32 & \textbf{77.28} & 84.85 & 74.56 & 85.94 & 86.73 & 87.17
      & +1.38 \\
      \midrule

      MC-Dr
      & 54.32 & 92.39 & 53.18 & 91.70 & 58.85 & 92.33 & 83.98 & 76.02 & 82.79 & 76.30 & 84.60 & 74.09 & 84.43 & 85.21 & 85.89
      & +0.05 \\
      \textbf{without LLR}
      & 57.09 & 93.82 & 55.24 & \textbf{93.07} & 61.84 & \textbf{93.53} & \textbf{85.02} & 76.13 & \textbf{83.73} & 77.07 & \textbf{85.39} & \textbf{75.51} & \textbf{86.33} & \textbf{86.89} & \textbf{87.38}
      & \textbf{+1.51} \\
      \textbf{with LLR}
      & \textbf{57.23} & \textbf{93.96} & \textbf{55.30} & 93.04 & \textbf{61.89} & \textbf{93.53} & 84.77 & \textbf{76.35} & 83.38 & \textbf{77.20} & 84.88 & 74.66 & 85.94 & 86.69 & 87.23
      & +1.38 \\
      \midrule

      TTAug
      & 56.95 & 95.20 & 56.74 & 94.39 & 62.09 & 95.67 & 86.43 & 75.20 & 85.57 & 75.02 & 85.09 & 75.43 & 83.02 & 83.80 & 84.30
      & +1.31 \\
      \textbf{without LLR}
      & 59.46 & \textbf{95.92} & \textbf{58.30} & \textbf{95.21} & 64.65 & \textbf{96.40} & \textbf{86.93} & 74.88 & \textbf{87.08} & 75.62 & \textbf{85.78} & \textbf{76.72} & \textbf{84.34} & \textbf{84.75} & \textbf{85.19}
      & \textbf{+2.39} \\
      \textbf{with LLR}
      & \textbf{60.00} & 95.77 & 58.08 & 94.34 & \textbf{65.19} & 95.38 & 84.91 & \textbf{75.76} & 83.90 & \textbf{76.00} & 85.59 & 75.96 & 83.52 & 84.32 & 84.77
      & +1.88 \\
      \midrule

      Tent
      & \textbf{52.87} & \textbf{94.92} & 53.83 & 92.17 & 60.02 & 93.23 & 84.19 & 74.88 & 83.09 & 75.98 & 84.70 & 73.93 & 85.23 & 86.32 & 86.71
      & +0.45 \\
      \textbf{without LLR}
      & 52.29 & 94.72 & \textbf{54.33} & \textbf{92.65} & \textbf{60.05} & \textbf{93.50} & \textbf{84.96} & 76.17 & \textbf{83.61} & 77.11 & \textbf{85.33} & \textbf{75.18} & \textbf{86.03} & \textbf{87.08} & \textbf{87.41}
      & \textbf{+1.01} \\
      \textbf{with LLR}
      & 52.34 & 94.77 & 54.23 & 92.54 & \textbf{60.05} & \textbf{93.50} & 84.67 & \textbf{76.54} & 83.31 & \textbf{77.28} & 84.81 & 74.27 & 85.81 & 86.94 & 87.32
      & +0.87 \\
      \midrule

      MEMO
      & 54.49 & 93.91 & 54.62 & 92.01 & 61.10 & 93.88 & 83.89 & 76.04 & 83.61 & 76.08 & 84.74 & 74.72 & 82.75 & 82.81 & 83.31
      & +0.18 \\
      \textbf{without LLR}
      & \textbf{56.19} & \textbf{95.22} & 55.95 & 93.00 & 61.63 & \textbf{93.97} & \textbf{84.95} & 76.30 & \textbf{83.68} & \textbf{76.92} & \textbf{85.33} & \textbf{75.40} & \textbf{83.62} & \textbf{83.74} & \textbf{84.23}
      & \textbf{+0.99} \\
      \textbf{with LLR}
      & \textbf{56.19} & 95.05 & \textbf{56.07} & \textbf{93.03} & \textbf{61.68} & 93.87 & 84.61 & \textbf{76.62} & 83.56 & 76.78 & 84.84 & 74.77 & 83.00 & 83.13 & 83.57
      & +0.76 \\
      \bottomrule
    \end{tabular}
  \end{sc}
  }
\end{table*}

Table~\ref{combined_vertical_no_horiz2} compares OATTA \emph{with} and \emph{without} the LLR gate across image, sensor, and language benchmarks, as add-ons to five baseline families (Base, MC-Dr, TTAug, Tent, MEMO). Overall, the ungated variant performs better on these naturally ordered streams: averaging the $\Delta_{\text{Base}}$ column across the five families, OATTA without LLR attains a mean improvement of +1.48\%, compared to +1.25\% with LLR. The drop is most pronounced when stacking on top of strong order-agnostic baselines; for example, on TTAug, without LLR reaches $\Delta_{\text{Base}}{=}$ +2.39\% whereas with LLR yields +1.88\%, consistent with the presence of exploitable temporal signal in real-world streams.

Despite this reduction, the LLR-gated variant remains a useful conservative default when temporal dependence is uncertain: even with LLR, OATTA achieves $\Delta_{\text{Base}}{=}$ +1.38\%, which exceeds the strongest standalone baseline (TTAug, $\Delta_{\text{Base}}{=}$ +1.31\%), while explicitly downweighting learned dynamics when they are not consistently supported.

\section{Controlled Data Streams Evaluation with and without LLR}
\label{app:llr_gate_eval_cont}

\begin{table*}[h]
\centering
\caption{\textbf{Main results on controlled streams (with LLR vs. without LLR (w/o LLR)).}
Accuracy (\%) is averaged over 10 runs. Baselines (Base/MC-Dr/TTAug/Tent/MEMO) are identical to the original Table~\ref{tab:accuracy_summary_extended} and omitted for compactness;
each entry reports accuracy after adding +Ours (w/o LLR) or +Ours with LLR (LLR) on top of the corresponding baseline.
Asterisks ($^*$) indicate statistically significant improvement ($p<0.05$) over the corresponding baseline by a two-sided Wilcoxon signed-rank test with Holm correction across the five baseline pairs (10 seeds).
LLR and w/o LLR robustly improve performance across diverse baselines and stream types.}
\label{tab:accuracy_oatta_vs_llr}
\setlength{\tabcolsep}{2.2pt}
\renewcommand{\arraystretch}{1.05}
\scriptsize
\resizebox{\textwidth}{!}{
\begin{tabular}{l|cc|cc|cc|cc|cc|cc}
\toprule
\multirow{2}{*}{\textbf{Stream}} &
\multicolumn{2}{c|}{\textbf{Base}} &
\multicolumn{2}{c|}{\textbf{MC-Dr}} &
\multicolumn{2}{c|}{\textbf{TTAug}} &
\multicolumn{2}{c|}{\textbf{Tent}} &
\multicolumn{2}{c|}{\textbf{MEMO}} &
\multicolumn{2}{c}{\textbf{$\Delta$ Avg}} \\
& \textbf{LLR} & \textbf{w/o LLR}
& \textbf{LLR} & \textbf{w/o LLR}
& \textbf{LLR} & \textbf{w/o LLR}
& \textbf{LLR} & \textbf{w/o LLR}
& \textbf{LLR} & \textbf{w/o LLR}
& \textbf{LLR} & \textbf{w/o LLR} \\
\midrule
S1: random
& \textbf{76.81} & 72.80
& \textbf{76.81} & 72.87
& \textbf{78.59} & 74.38
& \textbf{76.85} & 73.13
& \textbf{79.02} & 76.08
& \textbf{-0.15} $\pm$ 0.22 & -3.92 $\pm$ 0.57 \\
S2: sticky
& \sig{78.56} & \sig{\textbf{80.03}}
& \sig{78.55} & \sig{\textbf{80.02}}
& \sig{81.50} & \sig{\textbf{82.23}}
& \sig{78.27} & \sig{\textbf{79.87}}
& \sig{79.94} & \sig{\textbf{81.20}}
& 1.58 $\pm$ 0.23 & \textbf{2.89} $\pm$ 0.65 \\
\midrule
S2: sticky ($\alpha=0.10$)
& \textbf{76.89} & 72.84
& \textbf{76.86} & 72.79
& \textbf{78.99} & 72.78
& \textbf{77.13} & 73.20
& \textbf{79.27} & 76.53
& \textbf{-0.25} $\pm$ 0.21 & -4.45 $\pm$ 0.57 \\
S2: sticky ($\alpha=0.30$)
& \textbf{76.58} & 73.48
& \textbf{76.51} & 73.38
& \textbf{78.78} & 73.70
& \textbf{76.61} & 73.72
& \textbf{78.57} & 76.43
& \textbf{-0.00} $\pm$ 0.23 & -3.27 $\pm$ 0.58 \\
S2: sticky ($\alpha=0.50$)
& \sig{\textbf{77.61}} & 76.35
& \sig{\textbf{77.61}} & 76.22
& \sig{\textbf{80.15}} & 76.77
& \sig{\textbf{77.53}} & 76.46
& \sig{\textbf{79.37}} & 78.61
& \textbf{0.64} $\pm$ 0.25 & -0.94 $\pm$ 0.71 \\
S2: sticky ($\alpha=0.70$)
& \sig{78.56} & \sig{\textbf{80.03}}
& \sig{78.52} & \sig{\textbf{79.96}}
& \sig{\textbf{81.69}} & \sig{81.63}
& \sig{78.27} & \sig{\textbf{79.87}}
& \sig{80.00} & \sig{\textbf{81.27}}
& 1.57 $\pm$ 0.30 & \textbf{2.72} $\pm$ 0.65 \\
S2: sticky ($\alpha=0.85$)
& \sig{79.70} & \sig{\textbf{83.73}}
& \sig{79.72} & \sig{\textbf{83.73}}
& \sig{83.62} & \sig{\textbf{85.93}}
& \sig{79.10} & \sig{\textbf{83.60}}
& \sig{80.32} & \sig{\textbf{83.68}}
& 2.90 $\pm$ 0.28 & \textbf{6.54} $\pm$ 0.71 \\
S2: sticky ($\alpha=0.90$)
& \sig{80.62} & \sig{\textbf{85.44}}
& \sig{80.57} & \sig{\textbf{85.53}}
& \sig{85.56} & \sig{\textbf{88.23}}
& \sig{79.93} & \sig{\textbf{85.11}}
& \sig{81.17} & \sig{\textbf{84.93}}
& 3.73 $\pm$ 0.56 & \textbf{8.01} $\pm$ 0.58 \\
S2: sticky ($\alpha=0.95$)
& \sig{82.87} & \sig{\textbf{88.42}}
& \sig{82.92} & \sig{\textbf{88.46}}
& \sig{87.67} & \sig{\textbf{91.87}}
& \sig{82.03} & \sig{\textbf{88.06}}
& \sig{82.31} & \sig{\textbf{87.38}}
& 5.33 $\pm$ 0.49 & \textbf{10.61} $\pm$ 0.63 \\
S2: sticky ($\alpha=0.98$)
& \sig{83.48} & \sig{\textbf{89.67}}
& \sig{83.52} & \sig{\textbf{89.73}}
& \sig{88.70} & \sig{\textbf{93.53}}
& \sig{82.38} & \sig{\textbf{89.18}}
& \sig{82.24} & \sig{\textbf{87.62}}
& 6.35 $\pm$ 0.37 & \textbf{12.23} $\pm$ 0.75 \\
\midrule
S3: permuted
& \sig{78.14} & \sig{\textbf{78.89}}
& \sig{78.12} & \sig{\textbf{78.77}}
& \sig{80.69} & \sig{\textbf{80.85}}
& \sig{77.84} & \sig{\textbf{78.85}}
& \sig{79.65} & \sig{\textbf{80.76}}
& 1.17 $\pm$ 0.21 & \textbf{1.90} $\pm$ 0.82 \\
S4: regime switch
& \sig{78.14} & \sig{\textbf{78.39}}
& \sig{78.11} & \sig{\textbf{78.30}}
& \sig{\textbf{81.26}} & \sig{80.56}
& \sig{77.94} & \sig{\textbf{78.44}}
& \sig{79.86} & \sig{\textbf{80.34}}
& 1.05 $\pm$ 0.24 & \textbf{1.19} $\pm$ 0.67 \\
S5: three-phase
& \sig{78.27} & \sig{\textbf{78.98}}
& \sig{78.29} & \sig{\textbf{78.82}}
& \sig{\textbf{80.90}} & \sig{80.80}
& \sig{78.00} & \sig{\textbf{79.04}}
& \sig{79.86} & \sig{\textbf{80.61}}
& 1.06 $\pm$ 0.26 & \textbf{1.64} $\pm$ 0.33 \\
\bottomrule
\end{tabular}
}
\end{table*}

Table~\ref{tab:accuracy_oatta_vs_llr} provides a finer-grained view of when the LLR gate helps, beyond the summary in the main text. The clearest pattern is that the LLR primarily acts as a \emph{risk-control} mechanism in regimes where any learned transition prior is unreliable: on order-agnostic streams (S1: random) and under very weak temporal dependence ($\alpha\!\le\!0.30$), the ungated filter can overfit spurious transitions and incur large negative performance drift (e.g., $\Delta_{\text{Avg}}{=}$ -3.92\% on S1: random and -3.27\% at S2: sticky with $\alpha{=}0.30$), whereas the LLR keeps the average change near neutral (e.g., -0.15\% on S1: random and 0.00\% at S2: sticky with $\alpha{=}0.30$). This behavior is consistent across all five baseline families (Base, MC-Dr, TTAug, Tent, MEMO), indicating that the gate is not tied to a specific upstream predictor but rather to the presence (or absence) of exploitable temporal structure.

Conversely, once the stream becomes genuinely predictable, the same conservatism limits upside. As $\alpha$ increases, the ungated variant increasingly dominates, with the gap widening in the high-consistency regime (e.g., at $\alpha{=}0.95$, $\Delta_{\text{Avg}}{=}$ +10.61\% without LLR vs.\ +5.33\% with LLR; at $\alpha{=}0.98$, +12.23\% without LLR vs.\ +6.35\% with LLR). Notably, even for non-i.i.d.\ streams with partially disrupted order (S3: permuted, S4: regime switch, and S5: three-phase), the ungated variant tends to achieve slightly larger mean gains, suggesting that the filter can still exploit residual local consistency and recover after changes when it is not downweighted by the gate. Together, these results motivate using OATTA without LLR when a temporal signal is expected (to maximize adaptation) and enabling the LLR as a conservative default when the ordering may be weak, corrupted, or unknown.

\section{Statistical Significance Testing (Wilcoxon + Holm)}
\label{app:test}

For each stream and each baseline pair (e.g., Base vs.\ Base+OATTA), we obtain 10 paired accuracy measurements from 10 runs with different random seeds (identical stream instantiations/orderings are used for both methods, yielding paired observations). Because accuracy differences across seeds may be non-normal and we have a small sample size, we test whether the median paired accuracy difference is non-zero using a two-sided Wilcoxon signed-rank test~\citep{wilcoxon1945individual}. Since we test five baseline pairs within the same stream, we control the family-wise error rate (probability of any false positive within the stream) by applying Holm correction~\citep{holm1979simple} to the five $p$-values, and mark a result significant when the Holm-adjusted $p<0.05$.

\end{document}